\documentclass[final]{cvpr}

\usepackage{times}
\usepackage{epsfig}
\usepackage{graphicx}
\usepackage{amsmath}
\usepackage{amssymb}

\usepackage{xfrac}
\usepackage{subcaption}
\usepackage[export]{adjustbox}
\usepackage[table]{xcolor}
\usepackage{color}
\usepackage{rotating}
\usepackage{array}
\usepackage{fixltx2e}
\usepackage{xr}

\definecolor{myblue}{rgb}{.1,0.3,0.9}
\definecolor{rowblue}{RGB}{220,230,240}
\usepackage{booktabs}
\usepackage{tablefootnote}
\usepackage{colortbl}
\usepackage{tabularx}
\usepackage{multirow}
\usepackage[margin=0cm,font=small,labelfont=bf]{caption}
\usepackage{xr}

\usepackage[pagebackref=true,breaklinks=true,colorlinks,bookmarks=false]{hyperref}

\def\cvprPaperID{4729} 
\def\confYear{CVPR 2021}

\begin{document}

\title{Neural Scene Graphs for Dynamic Scenes}

\author{Julian Ost$^{1}$\qquad  Fahim Mannan$^{1}$\qquad Nils Th\"urey$^{2}$\qquad Julian Knodt$^{3}$\qquad Felix Heide$^{1, 3}$\vspace{5pt}\\
	$^1$Algolux  \quad $^2$Technical University of Munich \quad $^3$Princeton University \\ \\
	{\normalsize \url{http://light.princeton.edu/neural-scene-graphs}}}

\maketitle



\definecolor{Gray}{rgb}{0.5,0.5,0.5}
\definecolor{darkblue}{rgb}{0,0,0.7}
\definecolor{orange}{rgb}{1,.5,0} 
\definecolor{red}{rgb}{1,0,0} 

\definecolor{dai_ligth_grey}{RGB}{230,230,230}
\definecolor{dai_ligth_grey20K}{RGB}{200,200,200}
\definecolor{dai_ligth_grey40K}{RGB}{158,158,158}
\definecolor{dai_ligth_grey60K}{RGB}{112,112,112}
\definecolor{dai_ligth_grey80K}{RGB}{68,68,68}
\definecolor{dai_petrol}{RGB}{0,103,127}
\definecolor{dai_petrol20K}{RGB}{0,86,106}
\definecolor{dai_petrol40K}{RGB}{0,67,85}
\definecolor{dai_petrol80}{RGB}{0,122,147}
\definecolor{dai_petrol60}{RGB}{80,151,171}
\definecolor{dai_petrol40}{RGB}{121,174,191}
\definecolor{dai_petrol20}{RGB}{166,202,216}
\definecolor{dai_deepred}{RGB}{113,24,12}
\definecolor{dai_deepred20K}{RGB}{90,19,10}
\definecolor{dai_deepred40K}{RGB}{68,14,7}
\definecolor{rot}{RGB}{238, 28 35}
\definecolor{apfelgruen}{RGB}{140, 198, 62}
\definecolor{orange}{RGB}{244, 111, 33}
\definecolor{pink}{RGB}{237, 0, 140}
\definecolor{lila}{RGB}{128, 10, 145}
\definecolor{anthrazit}{RGB}{19, 31, 31}

\newcommand{\heading}[1]{\noindent\textbf{#1}}
\newcommand{\note}[1]{{\em{\textcolor{orange}{#1}}}}
\newcommand{\todo}[1]{{\textcolor{red}{\bf{TODO: #1}}}}
\newcommand{\comments}[1]{{\em{\textcolor{orange}{#1}}}}
\newcommand{\changed}[1]{#1}
\newcommand{\place}[1]{ \begin{itemize}\item\textcolor{darkblue}{#1}\end{itemize}}
\newcommand{\de}{\mathrm{d}}

\newcommand{\normlzd}[1]{{#1}^{\textrm{aligned}}}

\newcommand{\ttime}{\tau}               
\newcommand{\x}{\Vect{x}}               
\newcommand{\z}{z}               

\newcommand{\npixels}{n}               
\newcommand{\ntime}{t}               

\newcommand{\illfunc}     {g}
\newcommand{\pathfunc}     {s}
\newcommand{\camfunc}     {f}

\newcommand{\irradiance}{E}
\newcommand{\exposure}{b}
\newcommand{\pmdfunc}{f}                
\newcommand{\lightfunc}{g}              
\newcommand{\period}{T}                 
\newcommand{\freqm}{\omega}                
\newcommand{\illphase}{\rho}             
\newcommand{\sensphase}{\psi}             
\newcommand{\pmdphase}{\phi}            
\newcommand{\omphi}{{\omega,\phi}}      
\newcommand{\numperiod}{N}              
\newcommand{\att}{\alpha}               
\newcommand{\pathspace}{{\mathcal{P}}}  

\newcommand{\atan}{\operatorname{atan}}

\newcommand{\Fourier}{\mathfrak{{F}}}         
\newcommand{\conv}     {\otimes}
\newcommand{\corr}     {\star}
\newcommand{\Mat}[1]    {{\ensuremath{\mathbf{\uppercase{#1}}}}} 
\newcommand{\Vect}[1]   {{\ensuremath{\mathbf{\lowercase{#1}}}}} 
\newcommand{\Id}				{\mathbb{I}} 
\newcommand{\Diag}[1] 	{\operatorname{diag}\left({ #1 }\right)} 
\newcommand{\Opt}[1] 	  {{#1}_{\text{opt}}} 
\newcommand{\CC}[1]			{{#1}^{*}} 
\newcommand{\Op}[1]     {\Mat{#1}} 
\newcommand{\minimize}[1] {\underset{{#1}}{\operatorname{argmin}} \: \: } 
\newcommand{\maximize}[1] {\underset{{#1}}{\operatorname{argmax}} \: \: } 
\newcommand{\grad}      {\nabla}

\newcommand{\Basis}{\Mat{H}}         		
\newcommand{\Corr}{\Mat{C}}             
\newcommand{\correlem}{\bold{c}}             
\newcommand{\meas}{\Vect{b}}            
\newcommand{\Meas}{\Mat{B}}            
\newcommand{\MeasNormalized}{\Mat{B}^{\textrm{new}}}            
\newcommand{\Img}{H}                    
\newcommand{\img}{\Vect{h}}             
\newcommand{\latentresponse}{\alpha}

\newenvironment{customlegend}[1][]{%
        \begingroup
        \csname pgfplots@init@cleared@structures\endcsname
        \pgfplotsset{#1}%
    }{%
        \csname pgfplots@createlegend\endcsname
        \endgroup
    }%

    \def\addlegendimage{\csname pgfplots@addlegendimage\endcsname}

\begin{abstract}
Recent implicit neural rendering methods have demonstrated that it is possible to learn accurate view synthesis for complex scenes by predicting their volumetric density and color supervised solely by a set of RGB images. However, existing methods are restricted to learning efficient representations of static scenes that encode all scene objects into a single neural network, and lack the ability to represent dynamic scenes and decompositions into individual scene objects. In this work, we present the first neural rendering method that decomposes dynamic scenes into scene graphs. We propose a learned scene graph representation, which encodes object transformation and radiance, to efficiently render novel arrangements and views of the scene. To this end, we learn implicitly encoded scenes, combined with a jointly learned latent representation to describe objects with a single implicit function. We assess the proposed method on synthetic and real automotive data, validating that our approach learns dynamic scenes -- only by observing a video of this scene -- and allows for rendering novel photo-realistic views of novel scene compositions with unseen sets of objects at unseen poses. 
\end{abstract}

\vspace{-8pt}
\section{Introduction}
View synthesis and scene reconstruction from a set of captured images are fundamental problems
in computer graphics and computer vision. Classical methods rely on sequential reconstructions
and rendering pipelines that first recover a compact scene representation, such as a point-cloud
or textured mesh using structure from
motion~\cite{argarwal2011romeinaday,hartley2003mvgeometry,schoenberger2016SfMrevisited,
schoenberger2016mvs}, which is then used to render novel views using efficient direct or global
illumination rendering methods. These sequential pipelines also allow for learning hierarchical
scene representations~\cite{shum1999hierarchicalSfM}, representing dynamic
scenes~\cite{costeira1995multibodyfactorizationformotion,ozden2010multibodySfM}, and
efficiently rendering novel views~\cite{bianco2018performanceevalSfM}. However, traditional
pipelines struggle to capture highly view-dependent features
at discontinuities, or illumination-dependent reflectance of scene objects.


Recently, these challenges due to view-dependent effects have been tackled by neural rendering
methods. The most successful methods~\cite{mildenhall2020nerf,lombardi2019neuralvolumes} depart from
explicit scene representations such as meshes and estimated BRDF models, and instead learn fully
implicit representations that embed three dimensional scenes in functions, supervised by a sparse
set of images during the training. Specifically, implicit scene representation like Neural
Radiance Fields (NeRF) by Mildenhall et al.~\cite{mildenhall2020nerf} encode scene representations within the weights of a neural network that map 3D locations and viewing directions to a neural radiance field. The novel renderings from this representation improve on previous methods of discretized voxel grids~\cite{sitzmann2019deepvoxels}. %

However, although recent learned methods excel at view-dependent interpolation that traditional methods struggle with, they encode the entire scene representation into a single, static network that does not allow for hierarchical representations or dynamic scenes that are supported by traditional pipelines. Thus, existing neural rendering approaches assume that the training images stem from an underlying scene that does not change between views.
More recent approaches, such as NeRF-W \cite{martinbrualla2020nerfw}, attempt to improve on this
shortcoming by learning to ignore dynamic objects that cause occlusions within the static scene.
Specifically, NeRF-W incorporates an appearance embedding and a decomposition of transient and static elements via
uncertainty fields. 
However, this approach still relies on the consistency of a static scene to learn the underlying representation. 

In this work, we present a first method that is able to learn a representation for complex, dynamic multi-object scenes. 
Our method decomposes a scene into its static and
dynamic parts and learns their representations structured in a scene graph, defined through the corresponding tracking information  within the underlying scene and supervised by the frames of a video. The
proposed approach allows us to synthesize novel views of a scene, or render views for completely
unseen arrangements of dynamic objects. Furthermore, we show that the method can also be used for 3D object detection via inverse rendering.

Using automotive tracking data sets, our experiments confirm that our method is capable of representing scenes with highly dynamic objects. We assess the method by generating unseen views of novel scene compositions with unseen sets of objects at unseen poses. \\
Specifically, we make the following contributions:
\begin{itemize}
	\setlength\itemsep{.2em}
    \item We propose a novel neural rendering method that decomposes dynamic, multi-object scenes into a learned scene graph with decoupled object transformations and scene representations.
    \item We learn object representations for each scene graph node directly from a set of video frames and corresponding tracking data. We encode instances of a class of objects using a shared volumetric representation.
    \item We validate the method on simulated and experimental data, with labeled and generated tracking data by rendering novel unseen views and unseen dynamic scene arrangements of the represented scene.
	\item We demonstrate that the proposed method facilitates 3D object detection using inverse rendering.
\end{itemize}

\section{Related Work}
Recently, the combination of deep learning approaches and traditional rendering methods from computer graphics has enabled researchers to synthesize photo-realistic views for static scenes, using RGB images and corresponding poses of sparse views as input. We review related work in the areas of neural rendering, implicit scene representations, scene graphs, and latent object descriptors.

\vspace{0.5em}\noindent\textbf{Implicit Scene Representations and Neural Rendering}
A rapidly growing body of work has explored neural rendering methods~\cite{tewari2020NeuralSTAR}
for static scenes. Existing methods typically employ a learned scene representation and a
differentiable renderer. Departing from traditional representations from computer graphics,
which explicitly models every surface in a scene graph hierarchy, neural scene
representations implicitly represent scene features as outputs of a neural network. Such
scene representations can be learned by supervision using images, videos, point clouds, or a
combination of those. Existing methods have proposed to learn features on discrete geometric
primitives, such as points \cite{aliev2020neuralpointbased,pittaluga2019revealing}, multi-planes
\cite{flynn2019DeepView,lu2020layeredNeural,mildenhall2019llff,srinivasan2019viewextrapolationmultiplaneimages,zhou2018stereo},
meshes \cite{chen2018deepSLF,thies2019deferred} or higher-order primitives like voxel grids
\cite{lombardi2019neuralvolumes,sitzmann2019deepvoxels,yan2016perspective}, or implicitly
represent features
\cite{liu2020neuralsparsevoxel,mescheder2019occupancynetworks,niemeyer2020dvr,park2019deepsdf,xu2019disn,mildenhall2020nerf}
using functions $F: \mathbb{R}^{3}\to \mathbb{R}^{n}$ that map a point in the continuous 3D
space to a $n$-dimensional feature space as classified by the State of the Art report~\cite{tewari2020NeuralSTAR}. While traditional and explicit scene representations
are interpretable and allow for decomposition into individual objects, they do not scale
with scene complexity. Implicit representations model scenes as functions, typically encoding
the scene using a multilayer perceptron (MLP) and encode the scene in the weights of the MLP.
This gives up interpretability, but allows for a continuous view interpolation that is less
dependent on scene resolution~\cite{mildenhall2020nerf}. The proposed method closes this gap by introducing a hierarchical scene-graph model with object-level implicit representations.

Differentiable rendering functions have made it possible to learn scene
representations~\cite{mescheder2019occupancynetworks,niemeyer2020dvr,park2019deepsdf,mildenhall2020nerf}
from a set of images. Given a camera's extrinsic and intrinsic parameters, existing methods rely
on differentiable ray casting or ray marching through the scene volume to infer the features for
sampling points along a ray. Varying the extrinsic parameters of the camera at test time enables
novel view synthesis for a given static scene. Niemeyer et al.~\cite{niemeyer2020dvr} used such
an approach for learning an implicit scene representations for object shapes and
textures. Departing from strict shape, representations like Scene Representation Networks (SRNs)
\cite{sitzmann2019srns} and the approach of Mescheder et al.~\cite{mescheder2019occupancynetworks}, Mildenhall et al.~\cite{mildenhall2020nerf} output a color value conditioned on a ray's direction, allowing
NeRF~\cite{mildenhall2020nerf} to achieve high-quality, view-dependent effects. Our work builds
on this approach to model dynamic scene objects and we describe NeRF's concept in detail in the Supplemental Material. All methods described above rely on static scenes with consistent
mappings from a 3D point to a 2D observation. The proposed method lifts this assumption on
consistency, and tackles dynamic scenes by introducing a learned scene graph representation.
\begin{figure*}[t]
\vspace{-7mm}
	\begin{subfigure}{.5\textwidth}
		\centering
		\hspace{-3pt}
		\vspace{-12pt}
		\caption{Neural scene graph in isometric view.}
		\includegraphics[width=0.78\textwidth]{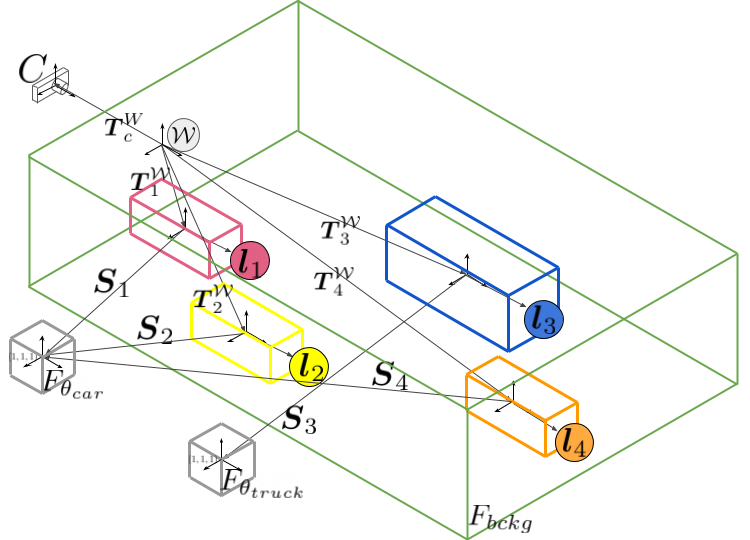}
		\label{fig:scene_graph_iso}
	\end{subfigure}
	\begin{subfigure}{.5\textwidth}
		\centering
		\caption{Neural scene graph from the ego-vehicle view.}
		\hspace{-20pt}
		\includegraphics[width=0.9\textwidth]{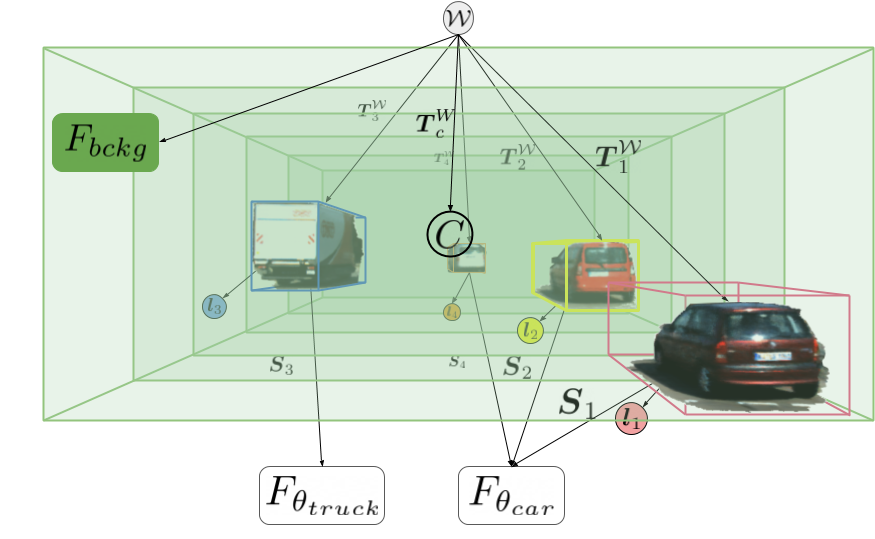}
		\label{fig:scene_graph_perspective}
	\end{subfigure}
  \vspace{-10pt}
\caption{(a) Three dimensional and (b) projected view of the same scene graph $\mathcal{S}_i$. The nodes are visualized as boxes with their local cartesian coordinate axis. Edges with transformations and scaling between the parent and child coordinate frames are visualized by arrows with their transformation and scaling matrix, $T^w_o$ and $S_o$. The corresponding latent descriptors are denoted by $\boldsymbol{l}_o$ and the representation nodes for the unit-scaled bounding boxes by $F_{\theta}$. No transformation is assigned to the background, which is already aligned with $\mathcal{W}.$}
\label{fig:scene_graphs}
\vspace{-15pt}
\end{figure*}

\vspace{0.5em}\noindent\textbf{Scene Graph Representations}
Scene graphs are traditional hierarchical representations of scenes. Introduced almost 30 years
ago~\cite{Wernecke1993TheIM,sowizral1998introductionJava3D,sowizral2000scenegraphs,cunningham2001lessonsfromSG},
they model a scene as a directed graph which represents objects as leaf nodes. These leaf nodes
are grouped hierarchically in the directed graph, with transformations, such as translation,
rotation and scaling, applied to the sub-graph along each edge, with each transformations defined in the local frame of its parent node. The global transformation for each leaf-node can be
efficiently calculated by stacking transformations from the global scene frame to the leaf node, and allows for reuse of objects and groups that are common to multiple children nodes. Since the 90s, scene graphs have been adopted in rendering and modeling pipelines, including Open Inventor \cite{Wernecke1993TheIM}, Java 3D \cite{sowizral1998introductionJava3D}, and VRML \cite{ames1997vrml}. Different geometry leaf-node representations have been proposed in the past, including individual geometry primitives, lists or volumetric shapes \cite{nadeau2000volumeSceneGraphs}. Aside from the geometric description, a leaf node can also represent individual properties of their parent nodes such as material and lighting. In this work, we revisit traditional scene graphs as a powerful hierarchical representation which we combine with learned implicit leaf-nodes. 

\vspace{0.5em}\noindent\textbf{Latent Class Encoding}
In order to better learn over distributions of shapes, a number of prior
works~\cite{mescheder2019occupancynetworks,niemeyer2020dvr,park2019deepsdf} propose to learn shape descriptors that generalize implicit scene representations across similar objects. 
By adding a latent vector $\boldsymbol{z}$ to the input 3D query point, similar objects can be modeled using the same network. Adding generalization across a class of objects can also be seen in other recent work that either use an explicit encoder network to predict a latent code from an image~\cite{mescheder2019occupancynetworks,niemeyer2020dvr}, or use a latent code to predict network weights directly using a meta-network approach~\cite{sitzmann2019srns}. Instead of using an encoder network, Park et al.~\cite{park2019deepsdf} propose to optimize latent descriptors for each object directly following Tan et al.~\cite{tan1995reducingDataDimensionality}. From a probabilistic view, the resulting latent code for each object follows the posterior $p_{\theta}(\boldsymbol{z}_i|X_i)$, given the
object's shape $X_i$. 
In this work, we rely on this probabilistic formulation to reason over dynamic objects of the
same class, allowing us to untangle global illumination effects conditioned on object position and type.\\\vspace{0.1\baselineskip}
Before presenting the proposed method for dynamic scenes, we refer the  reader to the Supplemental Material for a review of NeRF by Mildenhall et al.~\cite{mildenhall2020nerf} as a successful implicit representation for static scenes.


\section{Neural Scene Graphs}\label{ch:neuralscenegraph}
In this section, we introduce the neural scene graph, which allows us to model scenes hierarchically. The scene graph $\mathcal{S}$, illustrated in Fig.~\ref{fig:scene_graphs}, is composed of a camera, a static node and and a set of dynamic nodes which represent the dynamic components of the scene, including the object appearance, shape, and class.

\vspace{0.5em}\noindent\textbf{Graph Definition}
We define a scene uniquely using a directed acyclic graph, $\mathcal{S}$, given by
\begin{equation}
\mathcal{S}=\langle \mathcal{W}, C, F, L, E\rangle .
\label{eq:def_nsg}
\end{equation}
where $C$ is a leaf node representing the camera and its intrinsics $K\in\mathbb{R}^{3\times 3}$,
the leaf nodes $F = F_{\theta_{bckg}}\cup \{ F_{\theta_{c}} \}^{N_{class}}_{c=1}$ represents both static and dynamic representation models, $L =\{\boldsymbol{l}_o\}^{N_{obj}}_{o=1}$ are leaf nodes that assign latent object codes to each representation leaf node, and $E$ are edges that either represent affine transformations from $u$ to $v$ or property assignments, that is
\vspace{-2pt}
\begin{equation}
\begin{aligned}
& E= \{e_{u,v} \in \{ [\text{ }], \boldsymbol{M}^{u}_{v} \} \}, \\
& \text{with } \boldsymbol{M}^{u}_{v} = \begin{bmatrix}
\boldsymbol{R} & \boldsymbol{t} \\
\boldsymbol{0} & 1
\end{bmatrix}, \boldsymbol{R} \in \mathbb{R}^{3 \times 3}, \boldsymbol{t} \in \mathbb{R}^{3}.
\end{aligned}
\label{eq:edge_definition}
\end{equation}
For a given scene graph, poses and locations for all objects can be extracted. For all edges originating from the root node $\mathcal{W}$, we assign transformations between the global world space and the local object or camera space with $\boldsymbol{T}^{\mathcal{W}}_v$. 
Representation models are shared and therefore defined in a shared, unit scaled frame. To represent different scales of the same object type, we compute the non-uniform scaling $\boldsymbol{S}_o$, which is assigned at edge $e_{o,f}$ between an object node and its shared representation model $F$.

To retrieve an object's local transformation and position $\boldsymbol{p}_{o} = [x,z,y]_{o}$, we traverse from the root node $\mathcal{W}$, applying transformations until the desired object node $F_{\theta_{o}}$ is reached, eventually ending in the frame of reference defined in Eq.~\ref{eq:world2obj}.

%

\vspace{0.5em}\noindent\textbf{Representation Models}
For the model representation nodes, $F$, we follow Mildenhall et al.~\cite{mildenhall2020nerf}
and represent scene objects as augmented implicit neural radiance fields. 
In the following, we describe two augmented models for neural radiance fields which are also illustrated in Fig.~\ref{fig:model_arch}, which represent scenes as shown in Fig.~\ref{fig:scene_graphs}. 

\subsection{Background Node}
\begin{figure}[t!]
\centering
	\vspace{-12pt}
	\begin{minipage}{.42\textwidth}
		\centering
		\vspace{-5pt}
		\includegraphics[width=\textwidth, trim={35 200 175 50},clip]{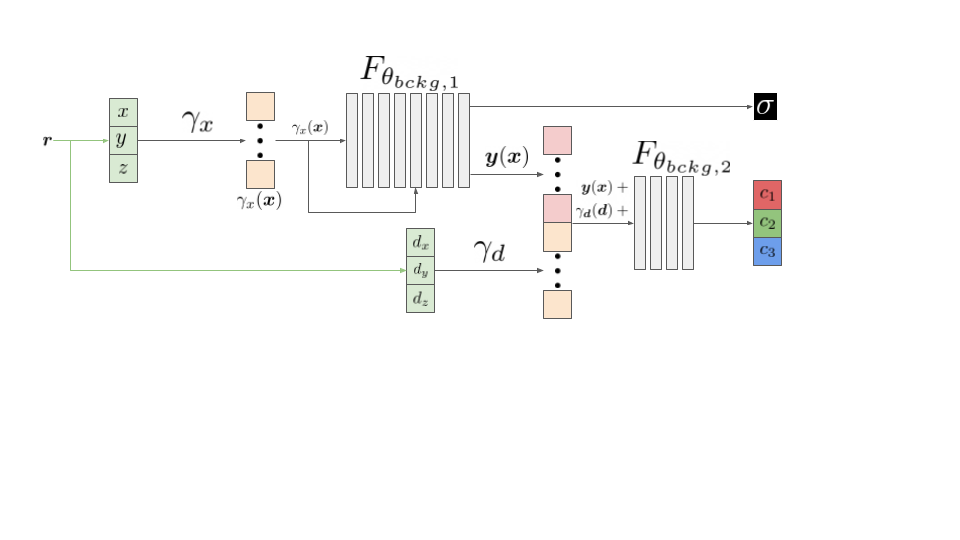}\\\vspace{-18pt}

		\hspace{-100pt}\text{Background Model}\\\vspace{4pt}
		\includegraphics[width=\textwidth, trim={110 100 147 40},clip]{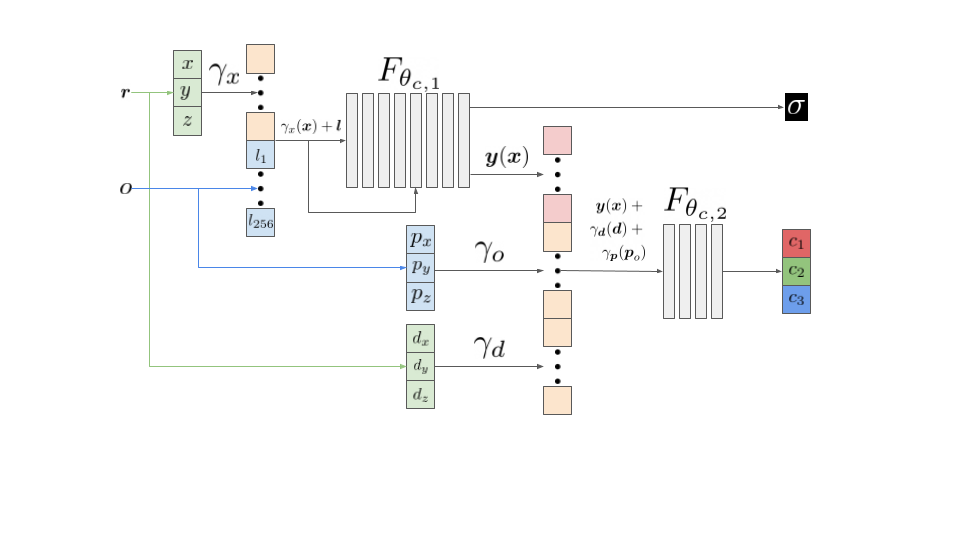}\\\vspace{-18pt}

		\hspace{-115pt}\text{Dynamic Model}\\\vspace{-4pt}
		\caption{Architectures of representation networks for static and dynamic models. Input point $\boldsymbol{x}=[x,y,z]$ and direction $\boldsymbol{d} = [d_x,d_y,d_z]$ from ray	$\boldsymbol{r}$, and for dynamic objects latent descriptor $\boldsymbol{l}_o$ and the pose of an object $o$ outputs $\sigma$ from the first stage and $\boldsymbol{c}$ from the second stage.}
		\label{fig:model_arch}
	\end{minipage}
	\vspace{-14pt}
\end{figure}

A single background representation model in the top of Fig.~\ref{fig:model_arch} approximates all static parts of a scene with a neural radiance fields, that, departing from prior work, lives on sparse planes instead of a volume.
The static background node function
$F_{\theta_{bckg}}:(\boldsymbol{x}, \boldsymbol{d}) \to (\boldsymbol{c},\boldsymbol{\sigma})$
maps a point $\boldsymbol{x}$ to its volumetric density and combined with a viewing direction to an emitted color.  %
Thus, the background representation is implicitly stored in the weights $\theta_{bckg}$.
We use a set of mapping functions, Fourier encodings~\cite{tancik2020fourier}, to aid learning
high-frequency functions in the MLP model. We map the positional and directional inputs with
$\gamma(\boldsymbol{x})$ and $\gamma(\boldsymbol{d})$ to higher-frequency feature vectors and 
pass those as input to the background model, resulting in the following two stages of the 
representation network, that is
\begin{align}
	\label{eq:MLP_bckg}
	[\sigma(\boldsymbol{x}), \boldsymbol{y}(\boldsymbol{x})] &= F_{\theta_{bckg,1}}(\gamma_{x}(\boldsymbol{x})) \\
	\boldsymbol{c}(\boldsymbol{x}) &= F_{\theta_{bckg,2}}(\gamma_{d}(\boldsymbol{d}), \boldsymbol{y}(\boldsymbol{x})) .
\end{align}

\vspace{-18pt}
\subsection{Dynamic Nodes}\label{ssec:dyn_nodes}
For dynamic scene objects, we consider each individual rigidly connected component of a scene
that changes its pose through the duration of the capture. A single, connected object is denoted
as a dynamic object $o$. Each object is represented by a neural radiance field in the local
space of its node and position $\boldsymbol{p}_{o}$. Objects with a similar appearance are combined in a class $c$ and share weights $\theta_{c}$ of the representation function $F_{\theta_{c}}$. A learned latent encoding vector $\boldsymbol{l}_o$ distinguishes the neural radiance fields of individual objects, representing a class of objects with
\begin{equation}
[\boldsymbol{c}(\boldsymbol{x}_o), \sigma(\boldsymbol{x}_o)] = F_{\theta_{c}} (\boldsymbol{l}_o, \boldsymbol{p}_{o}, \boldsymbol{x}_o, \boldsymbol{d}_o).
\end{equation}

\vspace{-4pt}
\vspace{0.5em}\noindent\textbf{Latent Class Encoding}
Training an individual representation for each object in a dynamic scene can easily lead to a large number of models and training effort. Instead, we aim to minimize the number of models that represent all objects in a scene, reason about shared object features and untangle global illumination effects from individual object radiance fields. Similar to Park et al. \cite{park2019deepsdf}, we introduce a latent vector $\boldsymbol{l}$ encoding an object's representation. Adding $\boldsymbol{l}_o$ to the input of a volumetric scene function $F_{\theta_{c}}$ can be thought as a mapping from the representation function of class $c$ to the radiance field of object $o$ as in
\vspace{-4pt}
\begin{equation}
\label{eq:class_to_object_function}
F_{\theta_{c}}(\boldsymbol{l}_{o}, \boldsymbol{x}, \boldsymbol{d}) = F_{\theta_{o}}(\boldsymbol{x}, \boldsymbol{d}).
\end{equation}
Conditioning on the latent code allows shared weights $\theta_{c}$ between all objects of class
$c$. Effects on the radiance fields due to global illumination only visible for some objects during training are shared across all objects of the same class.
We modify the input to the mapping function, conditioning the volume density on the global 3D location
$\boldsymbol{x}$ of the sampled point and a 256-dimensional latent vector $\boldsymbol{l}_{o}$, resulting in the following new first stage

\begin{equation}
\label{eq:MLP_dyn_1}
[\boldsymbol{y}(\boldsymbol{x}),\sigma(\boldsymbol{x})] = F_{\theta_{c,1}}(\gamma_{\boldsymbol{x}}(\boldsymbol{x}), \boldsymbol{l}_{o}).
\end{equation}

\begin{figure*}[htb]
	\centering
	\includegraphics[width=0.95\textwidth,trim={0 150 0 55}, clip]{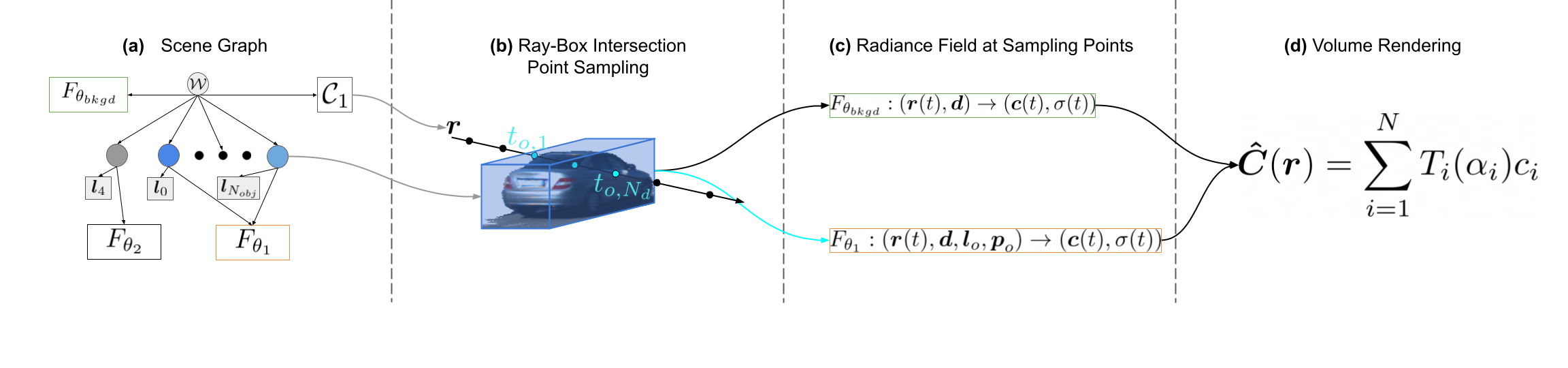}\vspace{-5pt}
	\caption{Overview of the Rendering Pipeline. (a) Scene graph modeling the scene (b) A sampled ray and the sampling points along the ray inside the background (black) and an object (blue) node (c) Mapping of each point to a color and density value with the corresponding representation function $F$ (d) Volume Rendering equation 
		over all sampling points along a ray from its origin to the far plane intersection} 
	\label{fig:rendering_pipeline}
	\vspace{-14pt}
\end{figure*}


\vspace{-4pt}
\vspace{0.5em}\noindent\textbf{Object Coordinate Frame}
The global location $\boldsymbol{p}_{o}$ of a dynamic object changes between frames, thus their radiance fields moves as well. We introduce local three-dimensional cartesian coordinate frames $\mathcal{F}_o$, fixed and aligned with an object pose. The transformation of a point in the the global frame $\mathcal{F}_{\mathcal{W}}$ to $\mathcal{F}_o$ is given by
\begin{equation}
\label{eq:world2obj}
\boldsymbol{x}_o = \boldsymbol{S}_o\boldsymbol{T}^w_o\boldsymbol{x}
\text{ with }\boldsymbol{x}_o \in [1,-1].
\end{equation}
scaling with the inverse length of the bounding box size $\boldsymbol{s}_{o} =
[L_o, H_o, W_o]$ with $\boldsymbol{S}_o = diag(1/\boldsymbol{s}_{o})$. This enables the models to learn size-independent similarities in $\theta_c$.

\vspace{0.5em}\noindent\textbf{Object Representation}
The continuous volumetric scene functions $F_{\theta_{c}}$ from Eq.~\ref{eq:class_to_object_function} are modeled with a MLP architecture presented in the bottom of Fig.~\ref{fig:model_arch}. The representation maps a latent vector $\boldsymbol{l}_o$, points $\boldsymbol{x} \in [-1, 1]$
and a viewing direction $\boldsymbol{d}$ in the local frame of $o$ to its
corresponding volumetric density $\boldsymbol{\sigma}$ and directional $\boldsymbol{c}$ emitted color. 
The appearance of a dynamic object depends on its interaction with the scene and global illumination which is changing for the object location $\boldsymbol{p}_o$. To consider the location-dependent effects, we add $\boldsymbol{p}_{o}$ in the global frame as another input.
Nevertheless, an object volumetric density $\sigma$ should not change based on on its pose in the
scene. To ensure the volumetric consistency, the pose is only considered for the emitted color and not the density. This adds the pose to the inputs $\boldsymbol{y}(t)$ and $\boldsymbol{d}$ of the second stage, that is
\begin{align}
\label{eq:MLP_dyn_2}
\boldsymbol{c}(\boldsymbol{x}, \boldsymbol{l}_o, \boldsymbol{p}_o) &= F_{\theta_{c,2}}(\gamma_{d}({\boldsymbol{d}}), \boldsymbol{y}(\boldsymbol{x}, \boldsymbol{l}_o), \boldsymbol{p}_{o}).
\end{align}
Concatenating $p_{o}$ with the viewing direction $\boldsymbol{d}$ preserves the view-consistent behavior from the original architecture and adds consistency across poses for the volumetric density. We again apply a Fourier feature mapping function to all low dimensional inputs $\boldsymbol{x}_o$, $\boldsymbol{d}_o$, $\boldsymbol{p}_{o}$. This results in the volumetric scene function for an object class $c$  
\begin{equation}
\label{eq:MLP_dyn}
F_{\theta_{c}}:(\boldsymbol{x}_o, \boldsymbol{d}_o, \boldsymbol{p}_{o}, \boldsymbol{l}_{o}) \to (\boldsymbol{c},\boldsymbol{\sigma});    \forall \boldsymbol{x}_o \in [-1,1].
\end{equation}



\section{Neural Scene Graph Rendering}\label{ch:rendering}
The proposed neural scene graph describes a dynamic scene and a camera view using a hierarchical structure. In this section, we show how this scene description can be used to render images of the scene, illustrated in
Fig.~\ref{fig:rendering_pipeline}, and how the representation networks at the leaf nodes are learned given a training set of images.

\subsection{Rendering Pipeline}\label{ssec:rendering_pipeline}
Images of the learned scene are rendered using a ray casting approach. The rays to be cast are
generated through the camera defined in scene $S$ at node $C$, its intrinsics $\boldsymbol{K}$, and the camera transformation
$\boldsymbol{T}^{\mathcal{W}}_c$. We model the camera $C$ with a pinhole camera model,
tracing a set of rays $\boldsymbol{r} = \boldsymbol{o}+ t\boldsymbol{d}$ through each pixel on
the film of size $H\times W$. Along this ray, we sample points at all graph nodes
intersected. At each point where a representation model is hit, a color and volumetric density
are computed, and we calculate the pixel color by applying a differentiable integration along the ray.
%

\vspace{0.5em}\noindent\textbf{Multi-Plane Sampling}
To increase efficiency, we limit the sampling at the static node to multiple planes as in \cite{zhou2018stereo}, resembling a 2.5 dimensional representation. We define $N_{s}$ planes, parallel to the
image plane of the initial camera pose $\boldsymbol{T}^w_{c,0}$ and equispaced between the near
clip $d_n$ and far clip $d_f$. For a ray, $\boldsymbol{r}$, we calculate the intersections
$\{t_i\}^{N_{s}}_{i=1}$ with each plane, instead of performing raymarching.

\vspace{0.5em}\noindent\textbf{Ray-box Intersection}
For each ray, we must predict color and density through each dynamic node through which it is traced. We check each ray from the camera $C$ for intersections with all dynamic nodes $F_{\theta_{o}}$, by translating the ray
to an object local frame and applying an efficient AABB-ray intersection test as proposed by
Majercik et al.~\cite{Majercik2018Voxel}. This computes all $m$ ray-box entrance and exit points
$(t_{o,1}, t_{o,N_{d}})$. For each pair of entrance and exit points, we sample $N_{d}$
equidistant quadrature points
\begin{equation}
	t_{o,n}=\frac{n-1}{N_{d}-1}(t_{o,N_{d}} - t_{o,1}) + t_{o,1},
\end{equation}
and sample from the model at $\boldsymbol{x}_o=\boldsymbol{r}(t_{o,n})$ and $\boldsymbol{d}_o$ with $n = \lbrack 1, N_d\rbrack $ for each ray. A small number of equidistant points $\boldsymbol{x}_o$ are enough to represent dynamic objects accurately while maintaining short rendering times.


\vspace{0.5em}\noindent\textbf{Volumetric Rendering}
Each ray $\boldsymbol{r}_j$ traced through the scene is discretized at $N_{d}$ sampling points at each of the $m_{j}$ dynamic node intersections and at $N_{s}$ planes, resulting in a set of quadrature points $\{ \{t_i\}^{N_{s}+m_{j} N_{d}}_{i=1} \}_j$.
%
The transmitted color $\boldsymbol{c}(\boldsymbol{r}(t_i))$ and volumetric density
$\sigma(\boldsymbol{r}(t_i))$ at each intersection point are predicted from the respective
radiance fields in the static node $F_{\theta_{bckg}}$ or dynamic node $F_{\theta_{c}}$. 
All sampling points and model outputs $(t,\sigma,\boldsymbol{c})$ are ordered along the ray
$\boldsymbol{r}(t)$ as an ordered set
\vspace{-8pt}
\begin{equation}
\{\{t_i, \sigma(\boldsymbol{x}_{j,i}), \boldsymbol{c}(\boldsymbol{x}_{j,i}) \}^{N_{s}+m_{j} N_{d}}_{i=1} |t_{i-1} < t_i \} .
\end{equation}
The pixel color is predicted using the rendering integral approximated with numerical quadrature as
\vspace{-4pt}
\begin{equation}
\begin{aligned}
\label{eqn:rendering_integral_composed_quadratur}
&\boldsymbol{\hat{C}}(\boldsymbol{r}) = \sum\limits_{i=1}^{N_{s}+m_jN_{d}} T_{i} \alpha_{i} \boldsymbol{c}_{i},
 \text{\:where \:}\\
& T_{i}=\mathrm{exp}(- \sum_{k=1}^{i-1} \sigma_{k} \delta_{k}) \text{\:\:and \:\:} \alpha_i = 1 - \mathrm{exp}(- \sigma_{i} \delta_{i}),
\end{aligned}
\end{equation}
where $\delta_{i}=t_{i+1}-t_{i}$ is the distance between adjacent points.
\begin{figure*}[t!]
\vspace{-6mm}
	\renewcommand{\arraystretch}{0.5}
	\centering
	\begin{tabular}[width=\textwidth]{ccc}
		(a) Reference&
		(b) Learned Object Nodes &
		(c) Learned Background \\

		\includegraphics[width=.3\textwidth, trim={10cm 0 0 0},clip]{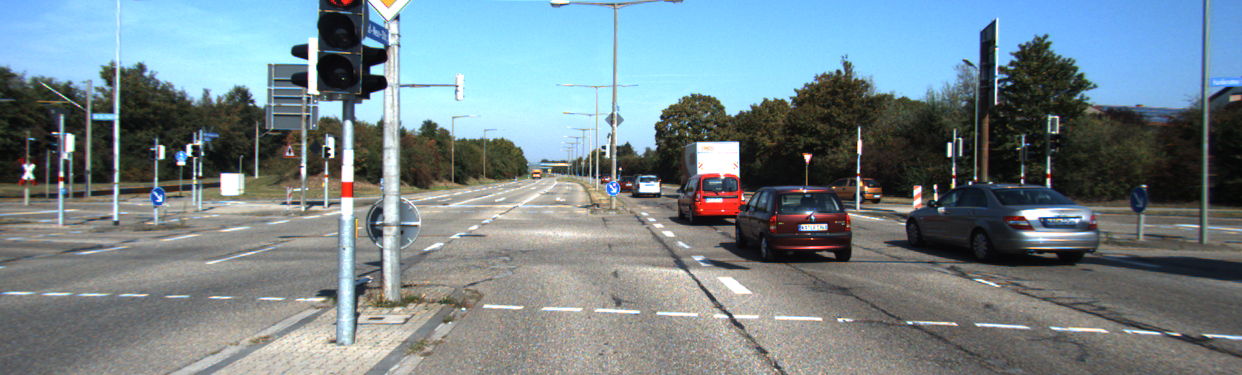}&
		\includegraphics[width=.3\textwidth, trim={10cm 0 0 0},clip]{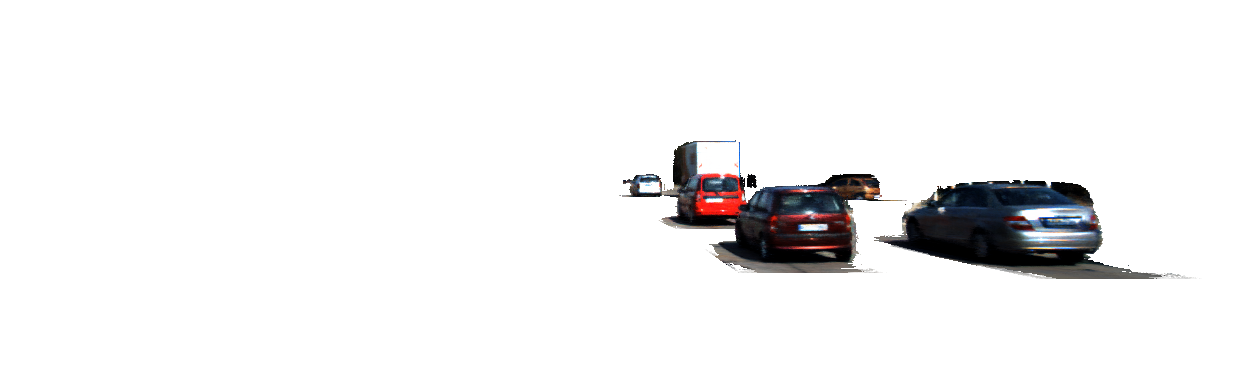}&
		\includegraphics[width=.3\textwidth, trim={10cm 0 0 0},clip]{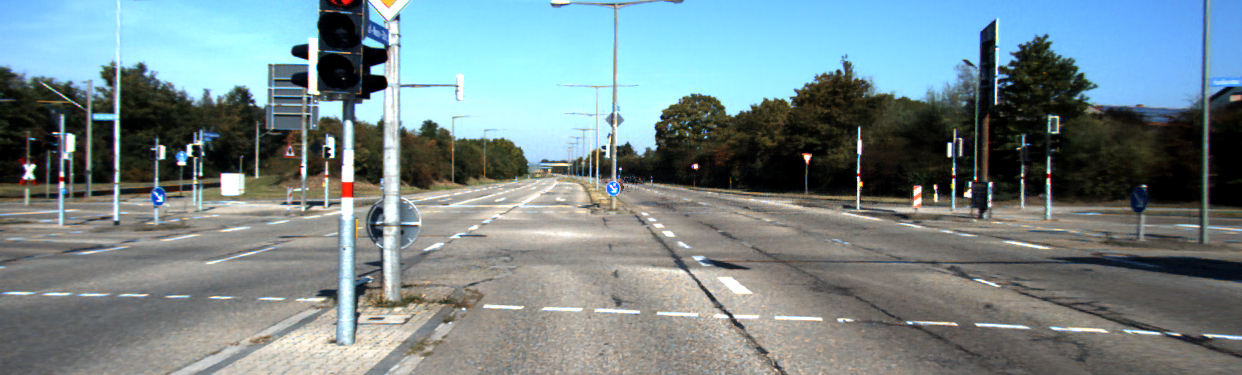} \\
		\vspace{0.2pt}

		(d) View Reconstruction &
		(e) Novel Scene &
		(f) Densely Populated Novel Scene \\

		\includegraphics[width=.3\textwidth,trim={10cm 0 0 0},clip]{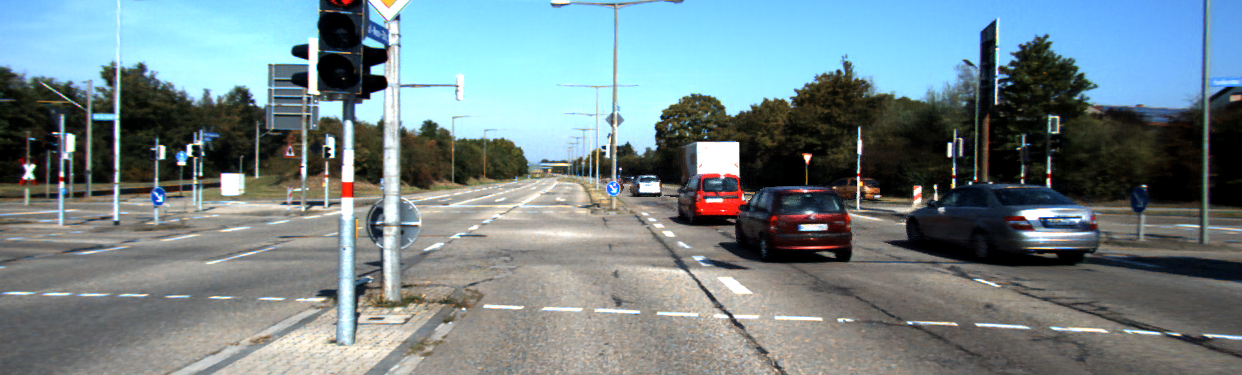}&
		\includegraphics[width=.3\textwidth, trim={10cm 0 0 0},clip]{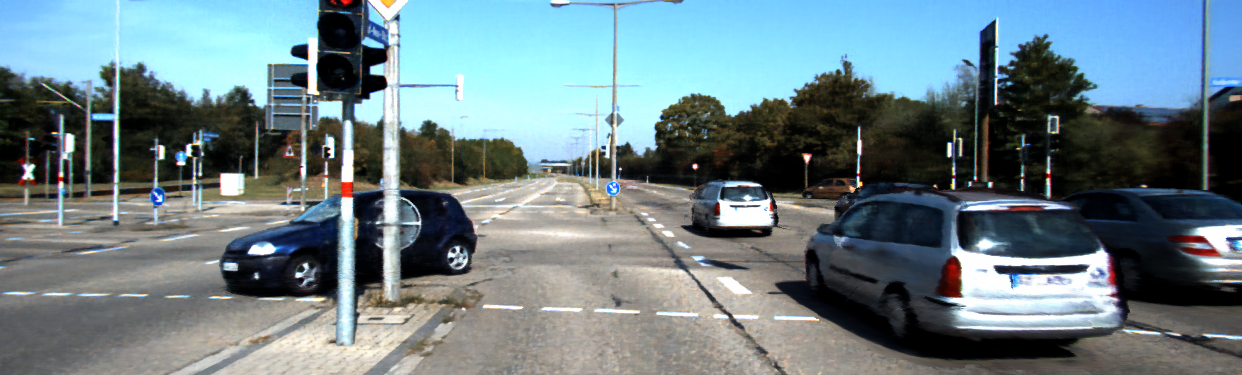} &
		\includegraphics[width=.3\textwidth,trim={10cm 0 0 0},clip]{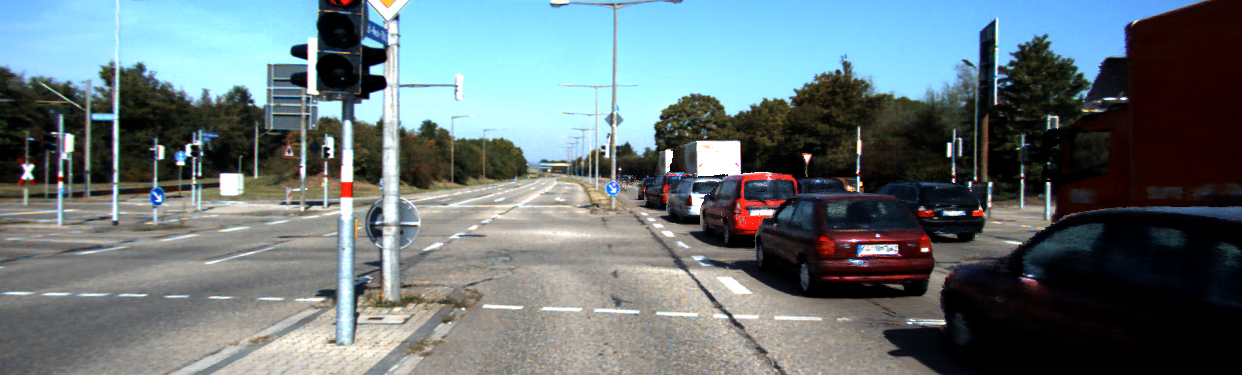} \\
	\end{tabular} \vspace{-8pt}
	\caption{Renderings from a neural scene graph on a dynamic
		KITTI~\cite{geiger2012kittivisonbenchmark} scene. The representations of all objects and the background are trained on images from a scene including (a). The method naturally decomposes
		the representations into the background (c) and multiple object representations (b). Using
    scene graphs from training renders reconstructions like (d). Novel scene renderings 
		 can come from randomly sampled nodes and translations into 
		 the drivable space, as displayed for medium and higher sampling densities  in (e) and (f).}
	\label{fig:result_decomposition}
	\vspace{-18pt}
\end{figure*}

\vspace{-8pt}
\subsection{Joint Scene Graph Learning}\label{sec:optimization}
For each dynamic scene, we optimize a set of representation networks at each node $F$. Our
training set consists of $N$ tuples $\{(\mathcal{I}_k, \mathcal{S}_k)\}^N_{k=1}$, the images
$\mathcal{I}_{k} \in \mathbb{R}^{H\times W\times 3}$ of the scene and the corresponding scene
graph $\mathcal{S}_k$. We sample rays for all camera nodes $\mathcal{C}_k$ at each pixel $j$. From given 3D tracking data, we take the transformations $\boldsymbol{M}^u_v$ to form the reference scene graph edges. Pixel values $\hat{C}_k,j$ are predicted for all $H \times W \times N$ rays $\boldsymbol{r}_k,j$.

We randomly sample a batch of rays $\mathcal{R}$ from all frames and associate each with
the respective scene graphs $\mathcal{S}_k$ and ground truth pixel value $C_k,j$.
We define the loss in Eq.~\ref{eqn:loss_f} as the total squared error between the predicted color $\hat{C}$ and the reference
values $C$. As in DeepSDF \cite{park2019deepsdf}, we assume a zero-mean multivariate Gaussian prior
distribution over latent codes $p(\boldsymbol{z}_o)$ with a spherical covariance
$\sigma^2\boldsymbol{I}$. For this latent representation, we apply a regularization to all
object descriptors with the weight $\sigma$.
\begin{equation}
\label{eqn:loss_f}
\mathcal{L} = \sum_{\boldsymbol{r}\in\mathcal{R}}\Vert \hat{C}(\boldsymbol{r}) - C(\boldsymbol{r}) \Vert_2^2 +\frac{1}{\sigma^2}\Vert\boldsymbol{z}\Vert^2_2
\end{equation}
At each step, the gradient to each trainable node in $L$ and $F$ intersecting with the rays in
$\boldsymbol{\mathcal{R}}$ is computed and back-propagated. The amount of nonzero gradients at a
node in a step depends on the amount of intersection points, leading to a varying amount of
evaluation points across representation nodes. We balance this ray sampling per batch. \\
We refer to the Supplemental Material for details on ray-plane and ray-box intersections and sampling.

\vspace{-6pt}
\section{Experiments}\label{ch:experiments}
\vspace{-5pt}
In this section, we validate the proposed neural scene graph method. To this end, we train neural scene graphs on videos from existing automotive datasets. We then modify the learned graphs to synthesize unseen frames of novel object arrangements, temporal variations, novel scene compositions and from novel views. We assess the quality with quantitative and qualitative comparisons against state-of-the-art implicit neural rendering methods.  The optimization for a scene takes about a full day or 350k iterations to converge on a single NVIDIA TITAN Xp GPU.

We choose to train our method on scenes from the KITTI
dataset~\cite{geiger2012kittivisonbenchmark}. For experiments on synthetic data, from KITTI's virtual counterpart, the Virtual KITTI 2 Dataset \cite{cabon2020vkitti2,geiger2012kittivisonbenchmark}, and for videos, we refer to the Supplementary Material and our project page.
Although the proposed method is not limited to automotive datasets, we use KITTI data as it has fueled research in object detection, tracking and scene understanding over the last decade. For each training scene, we optimize a neural scene graph with a node for each tracked object and one for the static background. The transformation matrices $\boldsymbol{T}$ and
$\boldsymbol{S}$ at the scene graph edges are computed from the tracking data.
We use $N_s=6$ planes and $N_d=7$ object sampling points. The object bounding box dimensions $\boldsymbol{s}_o$ are scaled to include the shadows of the object. The networks $\boldsymbol{F}_{\theta_{bckg}}$ and $\boldsymbol{F}_{\theta_{c}}$ have the architectures shown in Fig.~\ref{fig:model_arch}. We train the scene graph using the Adam optimizer~\cite{Kingma2015AdamAM} with a linear learning rate decay, see Supplemental Material for architecture and implementation details.
\begin{figure}[t!]
	\vspace{0pt}
	\renewcommand{\arraystretch}{0.5}
	\centering
	\begin{tabular}{@{}c@{}c@{}c@{}c@{}c@{}}
		\centering
		\includegraphics[width=\columnwidth/6, trim={25.5cm .5cm 6cm 7cm},clip]{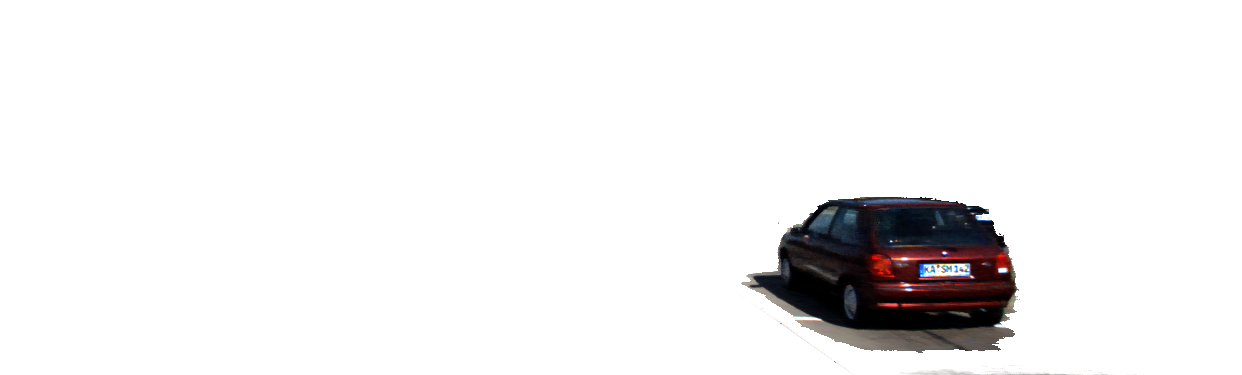}&
		\includegraphics[width=\columnwidth/6, trim={25.5cm .5cm 6cm 7cm},clip]{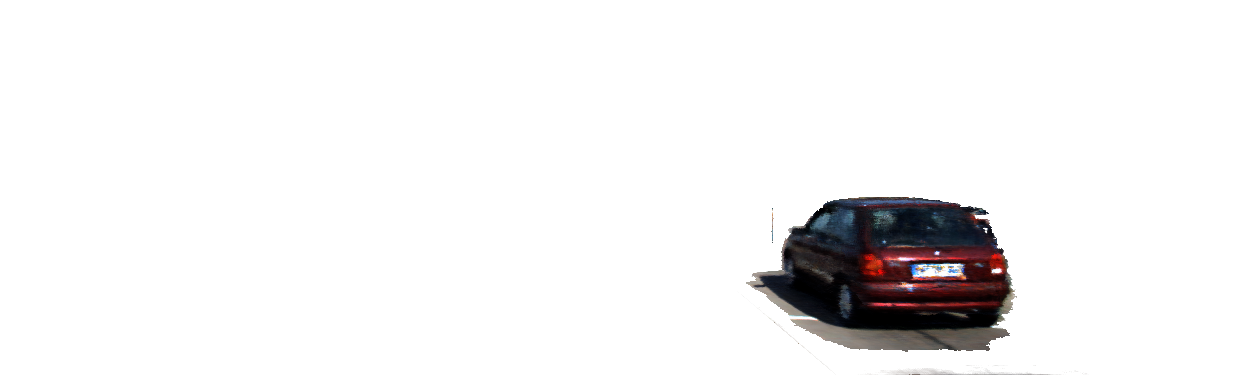}&
		\includegraphics[width=\columnwidth/6, trim={25.5cm .5cm 6cm 7cm},clip]{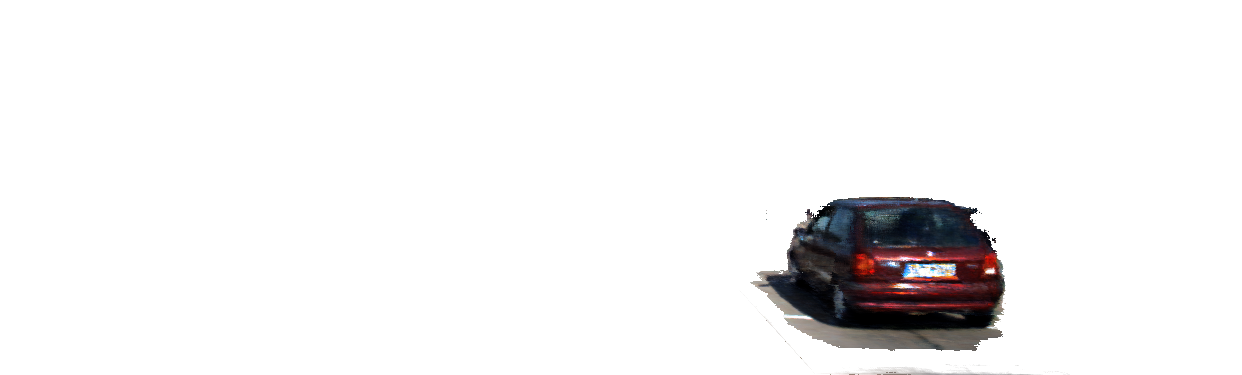}&
		\includegraphics[width=\columnwidth/6, trim={25.5cm .5cm 6cm 7cm},clip]{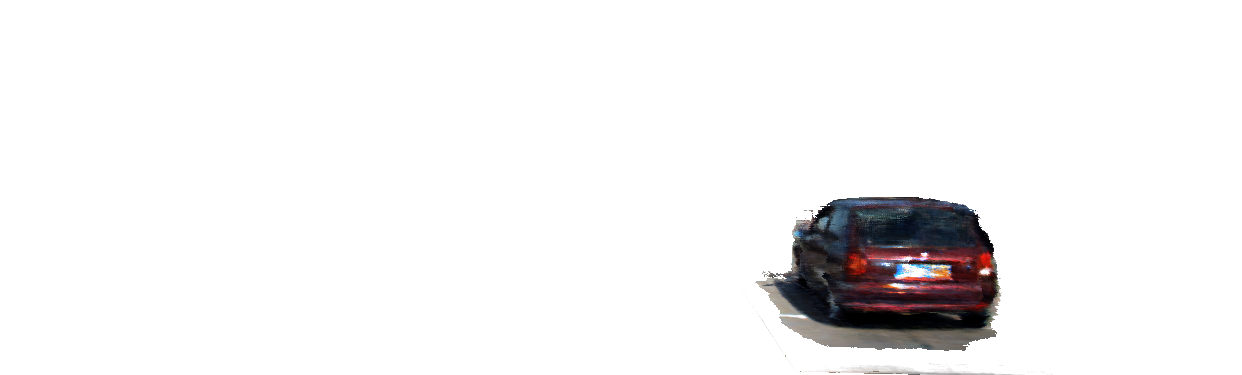}&
		\includegraphics[width=\columnwidth/6, trim={25.5cm .5cm 6cm 7cm},clip]{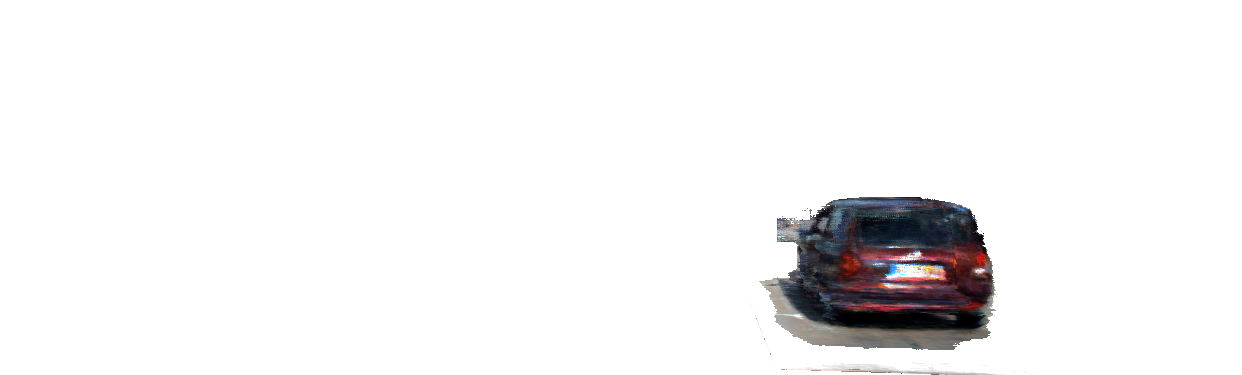}\\
		{\small Original Pose}&
		{\small 2.5$^\circ$}&
		{\small 5$^\circ$}&
		{\small 7.5$^\circ$}&
		{\small 10$^\circ$}\\
	\end{tabular}
	\vspace{-6pt}
	\caption{We synthesize multiple views of a learned scene graph leaf node by rotating yaw. The specular reflection on the trunk is maintained
		across rotations, demonstrating the ability of the method to learn view-variant lighting.}
	\label{fig:result_rotation}
	\vspace{-4pt}
\end{figure}
\begin{figure}[t!]
	\renewcommand{\arraystretch}{0.5}
	\centering
	\begin{tabular}{c@{}c@{}c@{}}
		\centering
		\hspace{-2pt}\includegraphics[width=.32\columnwidth,trim={24cm 1.7cm 3cm 5cm},clip]{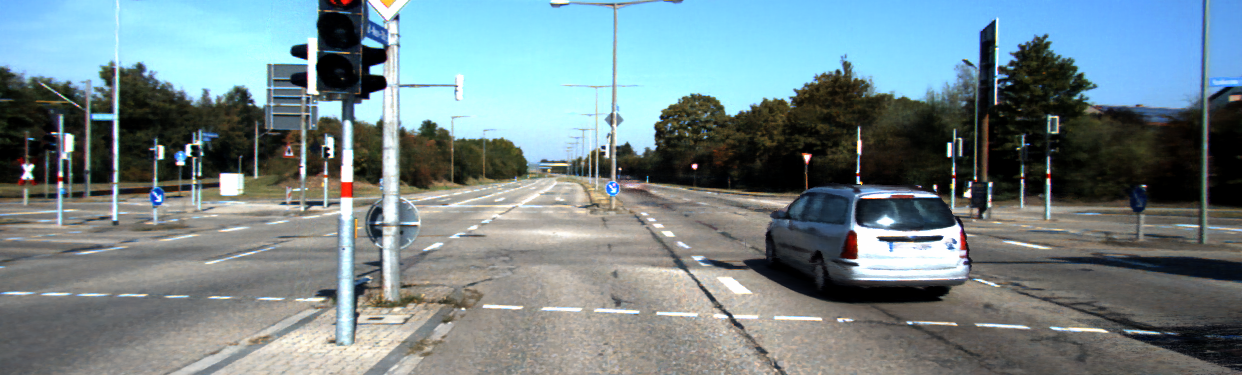} \hspace{1pt} & 
		\includegraphics[width=.32\columnwidth,trim={24cm 1.7cm 3cm 5cm},clip]{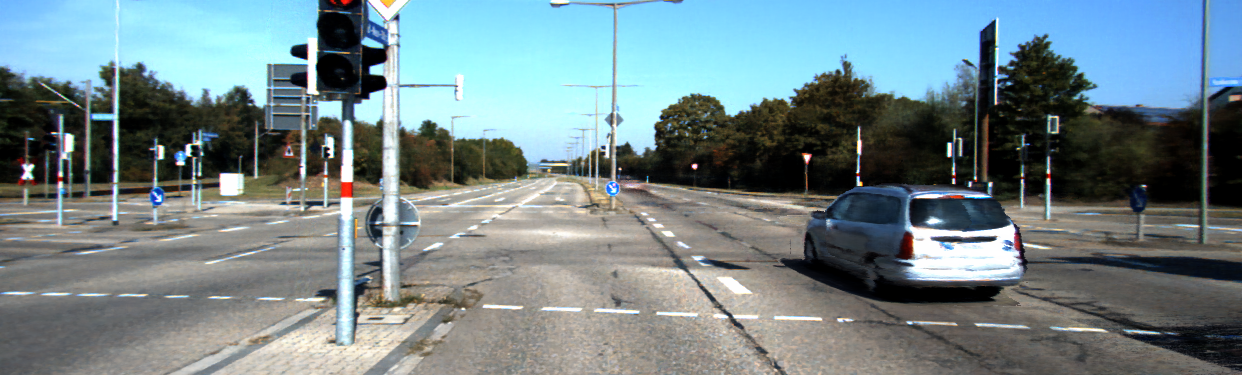} \hspace{1pt}& 
		\includegraphics[width=.32\columnwidth, trim={24cm 1.7cm 3cm 5cm},clip]{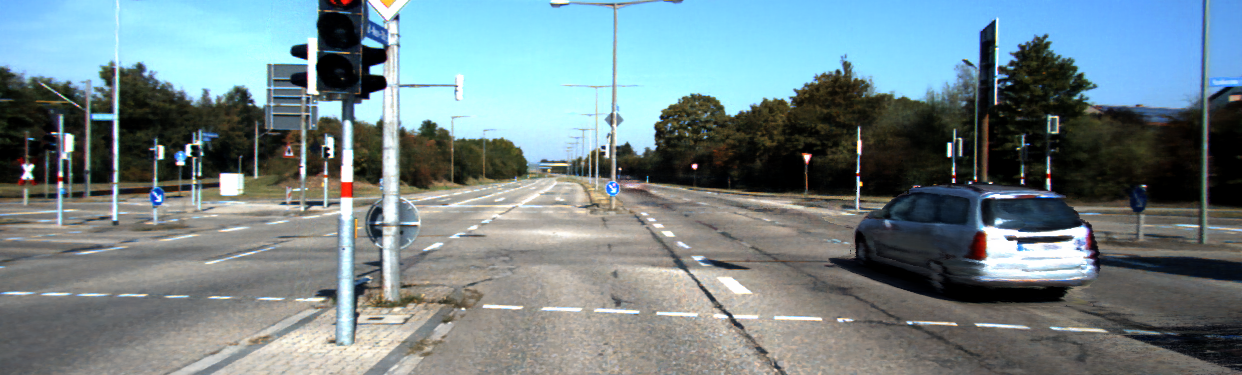}\\
		{\small Original Pose}&
		{\small 1~m shift}&
		{\small 2~m shift}\\
	\end{tabular}
	\vspace{-8pt}
	\caption{A learned vehicle leaf node is translated by 2 meters perpendicular to the movement observed during training. The model changes the shadows and specular highlights on the vehicle consistent with the learned global scene illumination.}
	\label{fig:result_translation}
	\vspace{-18pt}
\end{figure}

\begin{figure*}[htb]
	\vspace{-15pt}
	\renewcommand{\arraystretch}{0}
	\setlength{\tabcolsep}{2pt}
	\centering
	\begin{tabular}{ccc}
		\centering
		\includegraphics[width=.325\textwidth]{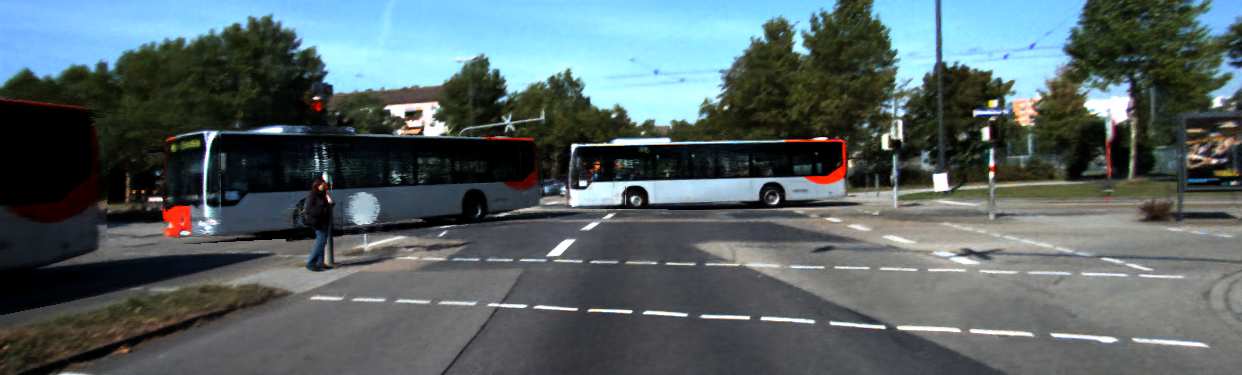} &
		\includegraphics[width=.325\textwidth]{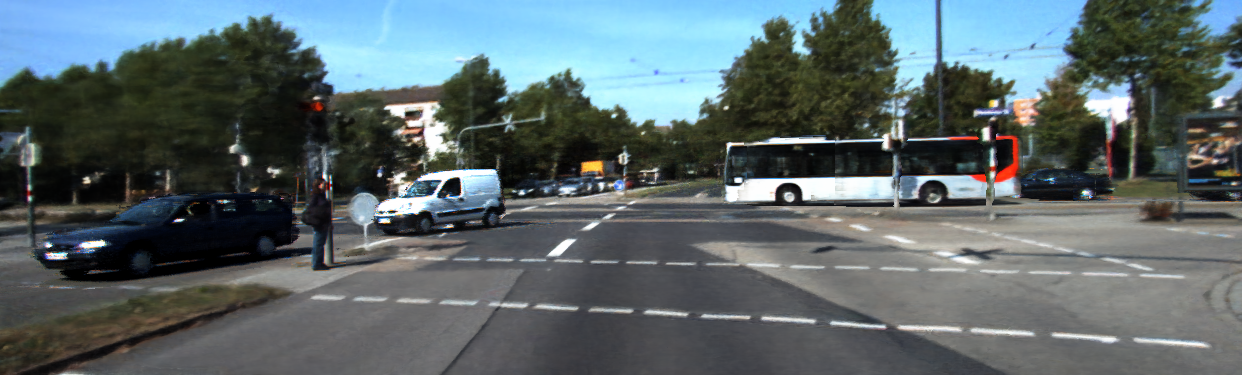} &
		\includegraphics[width=.325\textwidth]{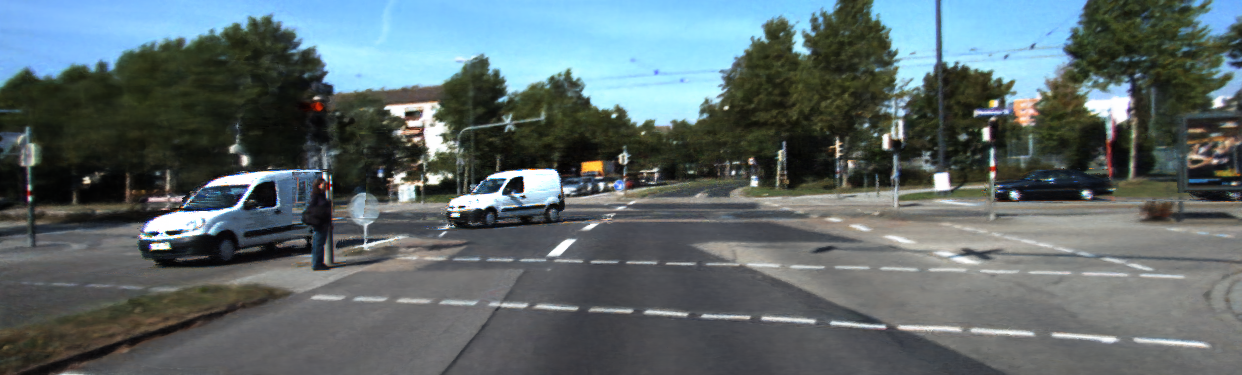} \\[3pt]
		\includegraphics[width=.325\textwidth]{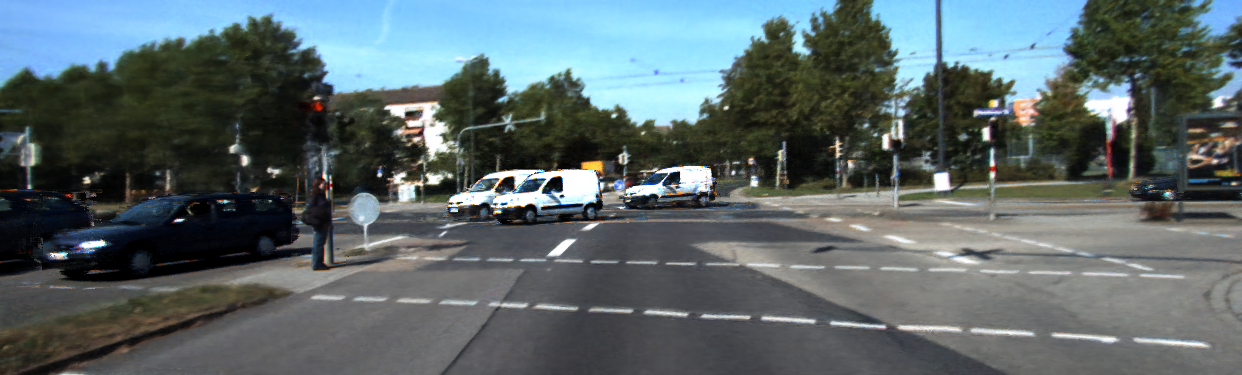} &
		\includegraphics[width=.325\textwidth]{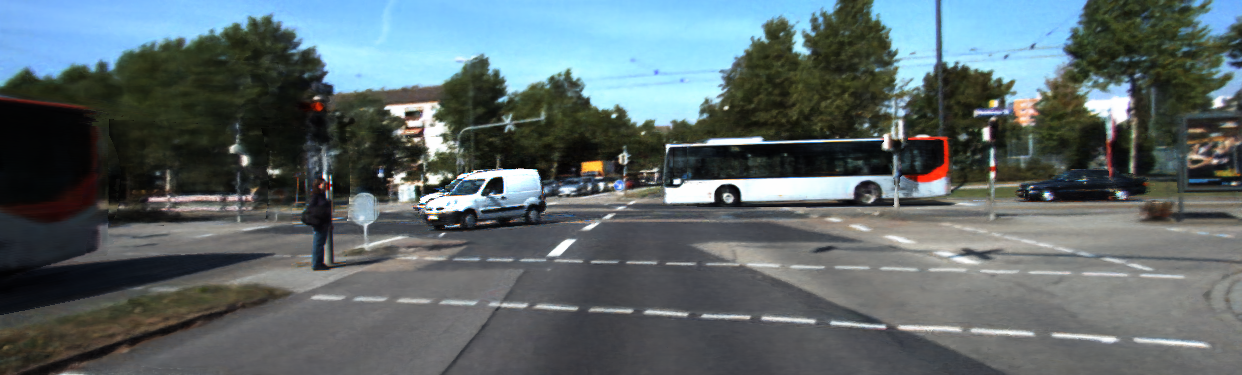} &
		\includegraphics[width=.325\textwidth]{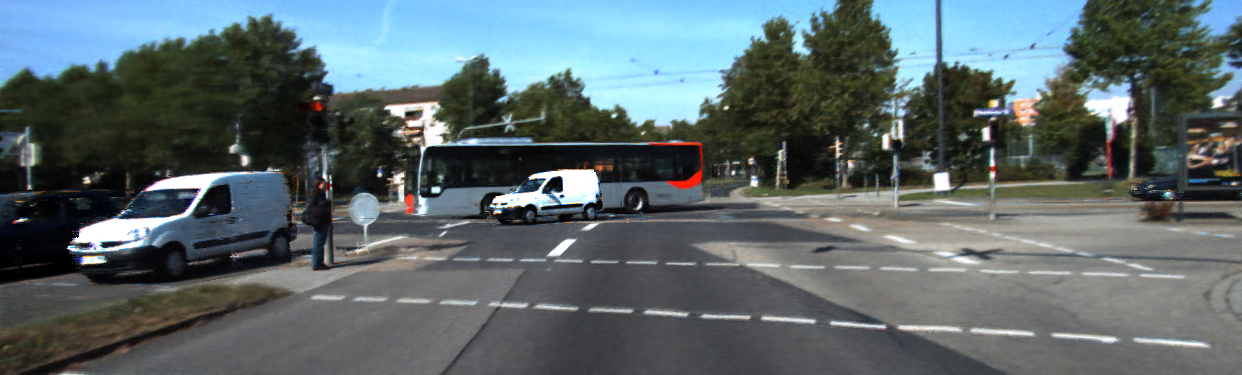} \\[-3pt]
	\end{tabular}
	\caption{Novel scene graph renderings. Rendered objects are learned from a subsequence of a
		scene from the KITTI data set~\cite{geiger2012kittivisonbenchmark}. The novel scene graphs are
		sampled from nodes and edges in the data set as well as new transformations sampled on the
		road lanes, not allowing for collisions between objects. Occlusion from the background and
    other objects can be observed in several frames.}
	\label{fig:result_novel_arrangement}
	\vspace{-18pt}
\end{figure*}

\begin{figure}[htb]
	\vspace{4pt}
	\centering
	\begin{subfigure}{.41\textwidth}
		\includegraphics[width=\textwidth]{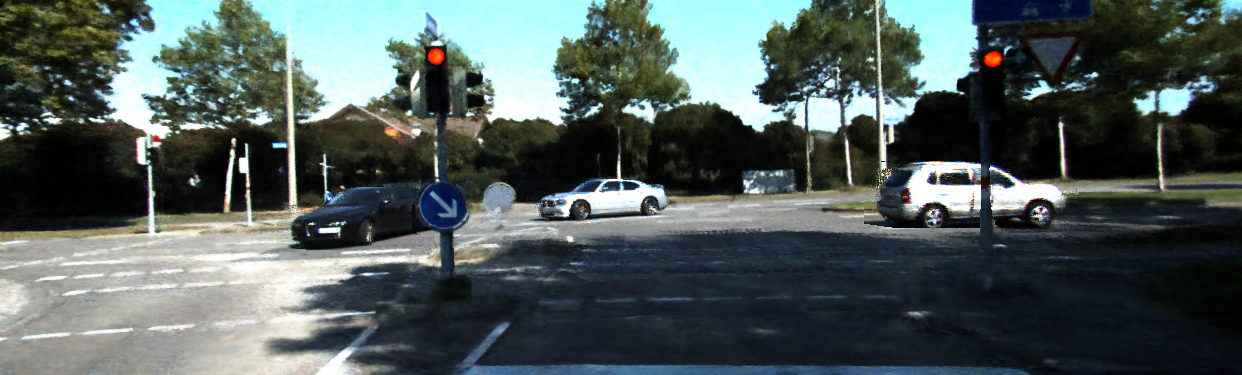}
		\vspace*{-35pt}
		\begin{center}
			{\footnotesize \textcolor{white}{Original Pose}}
		\end{center}
		\vspace{7.1pt}

	\end{subfigure}

	\vspace{-8pt}

	\begin{subfigure}{.41\textwidth}
		\includegraphics[width=\textwidth]{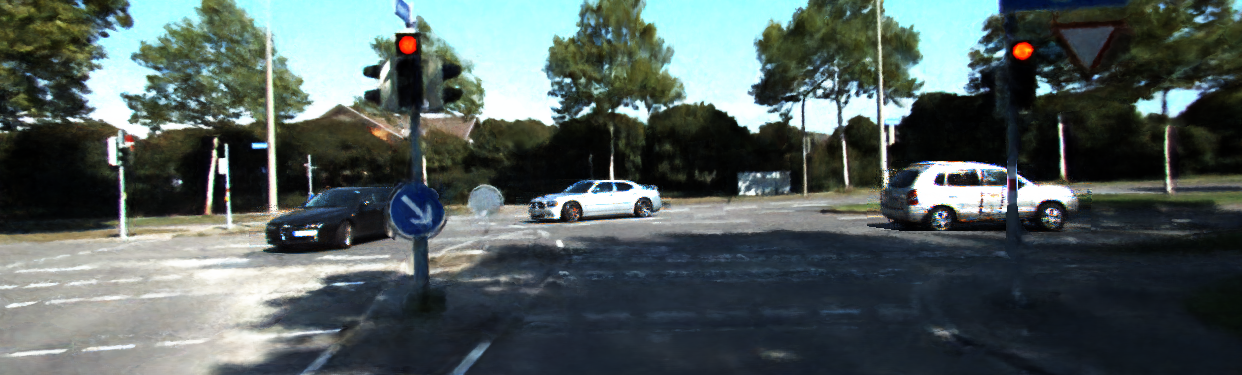}
		\vspace*{-35pt}
		\begin{center}
			{\footnotesize \textcolor{white}{1m forward}}
		\end{center}
		\vspace{7.1pt}
	\end{subfigure}

	\vspace{-8pt}

	\begin{subfigure}{.41\textwidth}
		\includegraphics[width=\textwidth]{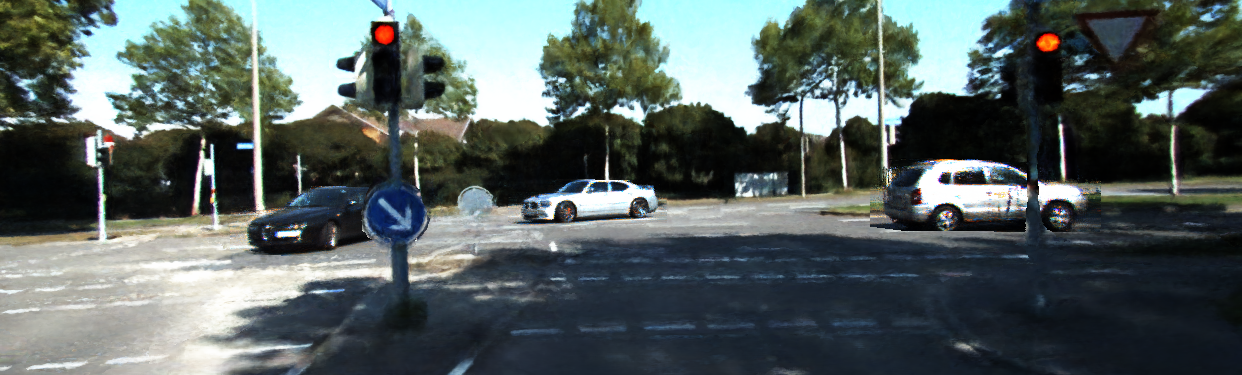}
		\vspace*{-35pt}
		\begin{center}
			{\footnotesize \textcolor{white}{2m forward}}
		\end{center}
	\end{subfigure}
	\caption{Unseen view synthesis results after translation. While the ego camera is moved by approximately 2~m into the scene, all other nodes of the scene graph are fixed. The method accurately handles occlusions from traffic lights and signs, highlighting the capabilities of the proposed scene graph for novel view synthesis.}
	\label{fig:cam_translation}

	\vspace{-18pt}
\end{figure}

\vspace{-3pt}
\subsection{Assessment}\label{ssec:experiment_results}
We present scene graph renderings for three scenes trained on dynamic
tracking sequences from KITTI~\cite{geiger2012kittivisonbenchmark}, captured using a stereo camera setup. Each training sequence consists of up to 90 time steps or 9 seconds and images of size $1242 \times 375$, each from two camera perspectives, and up to 12 unique, dynamic objects from different object classes. We assess possibilities to build the scene graph from tracking methods instead of dataset's annotations. \\

\vspace{-6pt}
\noindent\textbf{Foreground Background Decompositions}
Without additional supervision, the structure of the proposed scene graph model naturally
decomposes scenes into dynamic and static scene components. In
Fig.~\ref{fig:result_decomposition}, we present renderings of isolated nodes of a learned graph.
Specifically, in Fig.~\ref{fig:result_decomposition} (c), we remove all dynamic nodes and
render the image from a scene graph only consisting of the camera and background node. We
observe that the rendered node only captures the static scene components. Similarly, we
render dynamic parts in (b) from a graph excluding the static node. The shadow of each object is part of its dynamic representation and visible in the rendering. Aside from decoupling the background and all dynamic parts, the method accurately reconstructs (d) the target image of the scene (a). \\

\vspace{-6pt}
\noindent\textbf{Scene Graph Manipulations}
A learned neural scene graph can be manipulated at its edges and nodes. To demonstrate the flexibility of the proposed scene representation, we change the edge transformations of a learned node.
Fig.~\ref{fig:result_rotation} and~\ref{fig:result_translation} validate that these transformations preserve global light transport components such as reflections and shadows. The scene representation encodes global illumination cues implicitly through image color, a function of an object's location and viewing direction. Rotating a learned object node along its yaw axis in Fig.~\ref{fig:result_rotation} validates that the specularity on the car trunk moves in accordance to a fixed scene illumination, and retains a highlight with respect to the viewing direction.
In Fig.~\ref{fig:result_translation}, an object is translated away from its original position in the training set. In contrast to simply copying pixels, the model learns to correctly translate specular highlights after moving the vehicle.\\ 

\vspace{-4pt}
\noindent\textbf{Novel Scene Graph Compositions and View Synthesis}
In addition to pose manipulation and node removal from a learned scene graph, our method allows for constructing completely novel scene graphs, and novel view synthesis. 
Fig.~\ref{fig:result_novel_arrangement} shows view synthesis results for novel scene graphs that
were generated using randomly sampled objects and transformations. We constrain samples to the
joint set of all road trajectories that were observed, which we define as the driveable space. Note that these translations and arrangements have not been observed during training. \\
Similar to prior neural rendering methods, the proposed method allows for novel view synthesis after learning the scene representation. Fig.~\ref{fig:cam_translation} reports novel views for ego-vehicle motion into the scene, where the ego-vehicle is driving $\approx$ 2~m forward. We refer to the Supplemental Material for additional view synthesis results.

\begin{figure}[h!]
	\vspace{-2pt}
	\renewcommand{\arraystretch}{0.5}
	\centering
	\begin{tabular}{@{}c@{}c@{}c@{}}
		\centering
		\hspace{-2pt} \includegraphics[width=0.31\columnwidth, trim=400 0 0 100, clip]{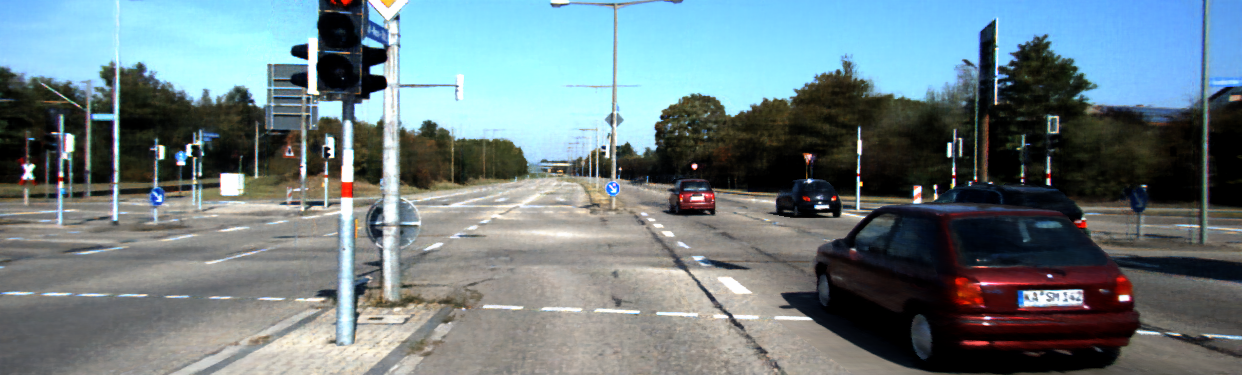} \hspace{1pt} &
		\includegraphics[width=0.31\columnwidth, trim=400 0 0 100, clip]{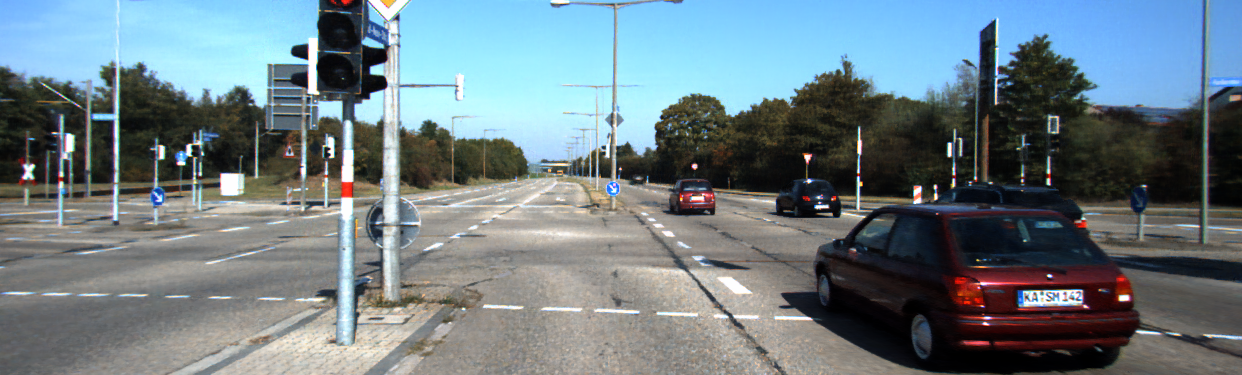} \hspace{1pt} &
		\includegraphics[width=0.31\columnwidth, trim=400 0 0 100, clip]{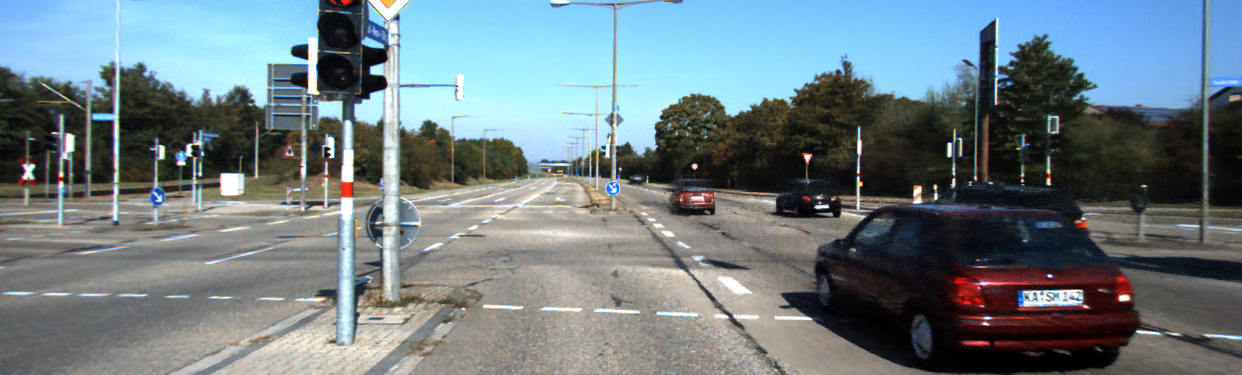} \\
		
		KITTI &
		PointRCNN + &
		CenterTrack \\
		
		GT-Tracking &
		AB3DMOT &
		\space \\
	\end{tabular}
	\vspace{-6pt}
	\caption{Comparison of neural scene graph renderings learned from labeled tracking data and tracking results obtained from off-the-shelf trackers. "PointRCNN + AB3DMOT" is a method that employs lidar 3D object detection and 3D tracking, and the corresponding scene graph achieves rendering quality comparable to scenes trained from manually annotated objects. "CenterTrack" is a camera-only method, see text, and still allows for robust results for close objects while degrading for objects at long distances.}
	\label{fig:tracking}
\end{figure}

\noindent\textbf{Scene Graphs from Noisy Tracker Outputs}
Neural scene graphs can be learned from manual annotations or the outputs of tracking methods. While temporally filtered human labels typically exhibit low labeling noise~\cite{geiger2012kittivisonbenchmark}, manual annotation can be prohibitively costly and time-consuming. We show that neural scene graphs learned from the output of existing tracking methods offer a representation quality comparable to the ones trained from manual annotations. Specifically, we compare renderings from scene graphs learned from KITTI~\cite{geiger2012kittivisonbenchmark} annotations and the following two tracking methods: we combine PointRCNN~\cite{shi2019pointrcnn}, a lidar point cloud object detection method, with the 3D tracker AB3DMOT~\cite{weng2020AB3DMOT}, and we evaluate on labels from a camera-only tracker CenterTrack~\cite{zhou2020tracking}, i.e., resulting in a method solely relying on unannotated video frames. Fig.~\ref{fig:tracking} shows qualitative results that indicate comparable neural scene graph quality when using different trackers, see also Supplemental Material. The monocular camera-only approach unsurprisingly degrades for objects at long distances, where the detection and tracking stack can no longer provide accurate depth information.

\begin{figure*}[t!]
	\vspace{-18pt}
	\hspace*{-10pt}
	\centering
	\renewcommand{\arraystretch}{0.5}
	\begin{tabular}[htb]{ll|cccc}
		\space & Reference & SRN & NeRF & NeRF + Time & Ours \\
		\rotatebox{90}{\parbox{1cm}{Recon-struction}} \hspace{-10pt} &
		\includegraphics[width=\textwidth/6, trim={400 0 0 0},clip]{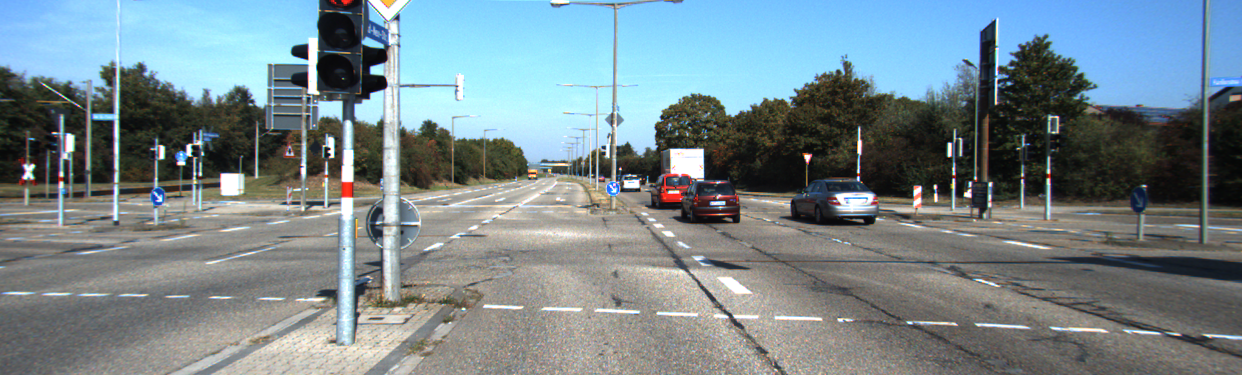}&
		
		\includegraphics[width=\textwidth/6, trim={0 100 0 108},clip]{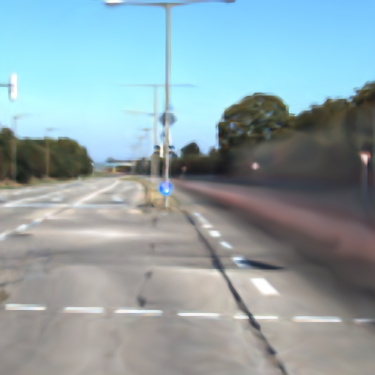}&
		\includegraphics[width=\textwidth/6, trim={400 0 0 0},clip]{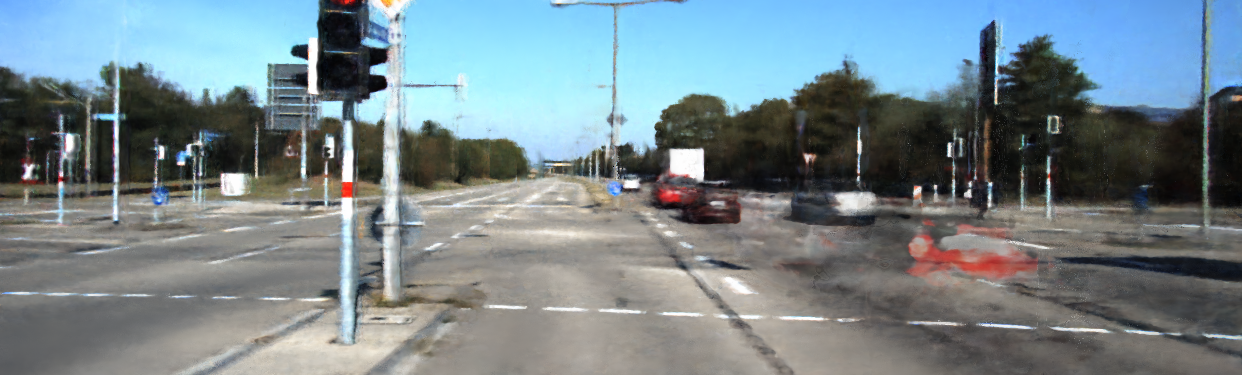}&
		\includegraphics[width=\textwidth/6, trim={400 0 0 0},clip]{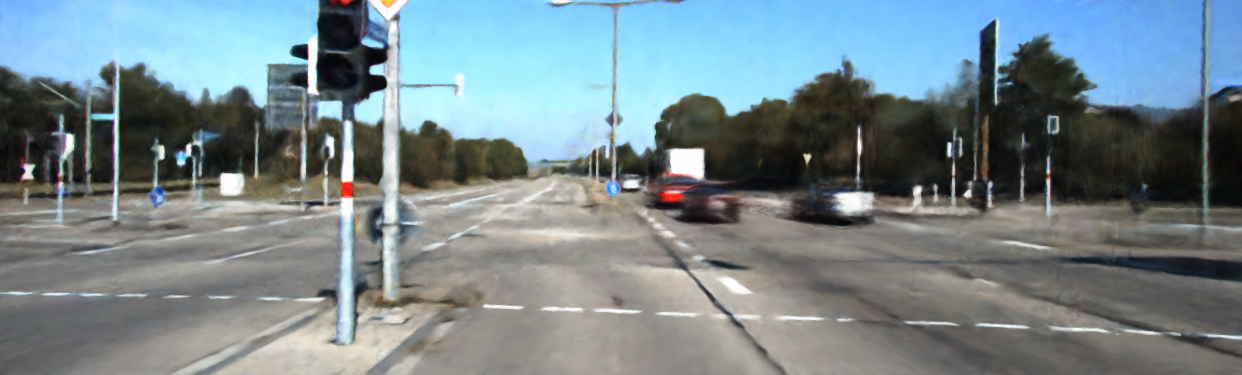}&
		\includegraphics[width=\textwidth/6, trim={400 0 0 0},clip]{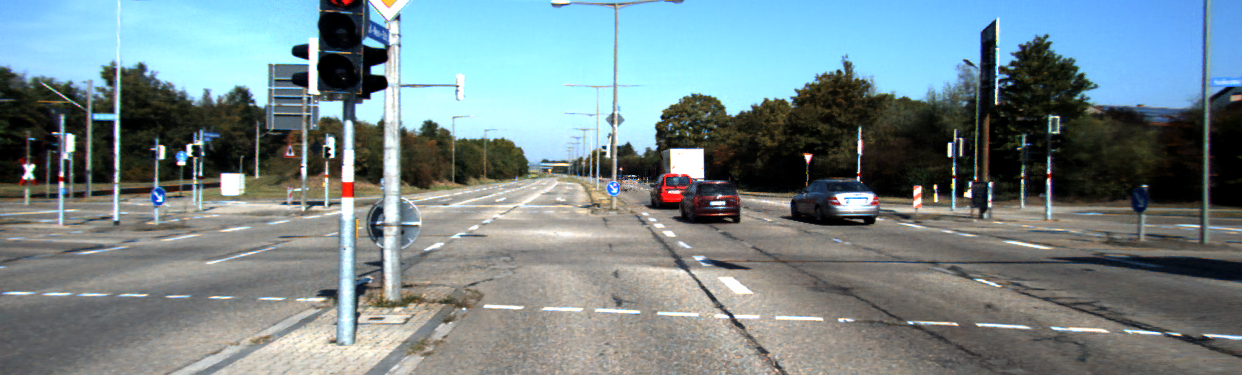}\\
		
		\rotatebox{90}{\parbox{1cm}{Novel \\ Scene}} \hspace{-10pt} &
		\includegraphics[width=\textwidth/6, trim={400 0 0 0},clip]{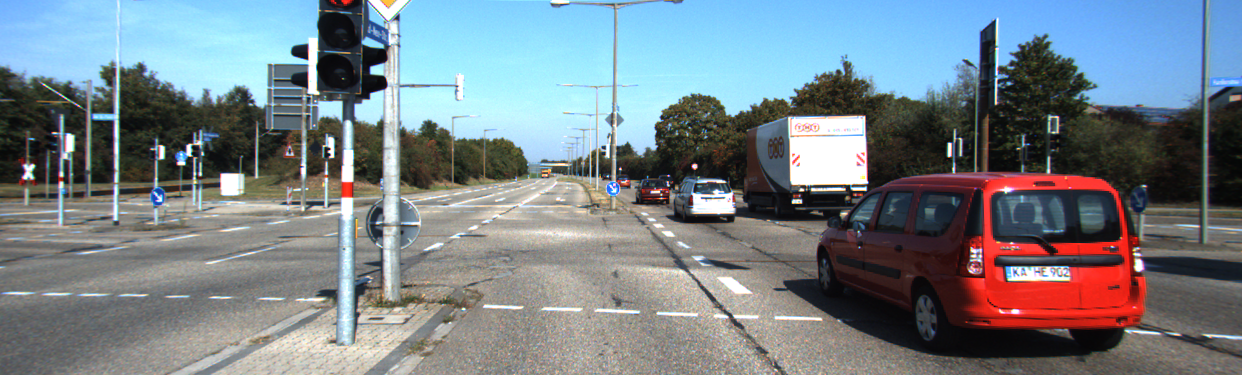}&
		
		\includegraphics[width=\textwidth/6, trim={0 100 0 108},clip]{fig/results/comparison/scene06_053_srn.png}&
		\includegraphics[width=\textwidth/6, trim={400 0 0 0},clip]{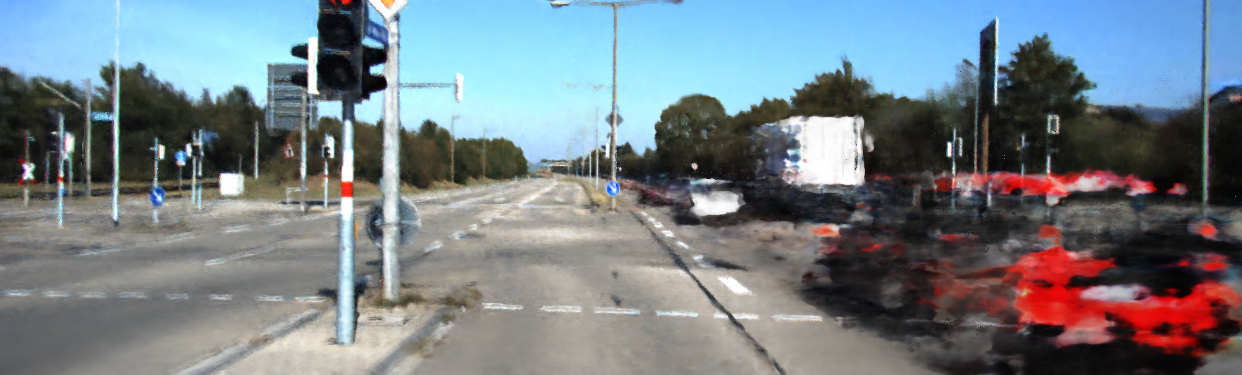}&
		\includegraphics[width=\textwidth/6, trim={400 0 0 0},clip]{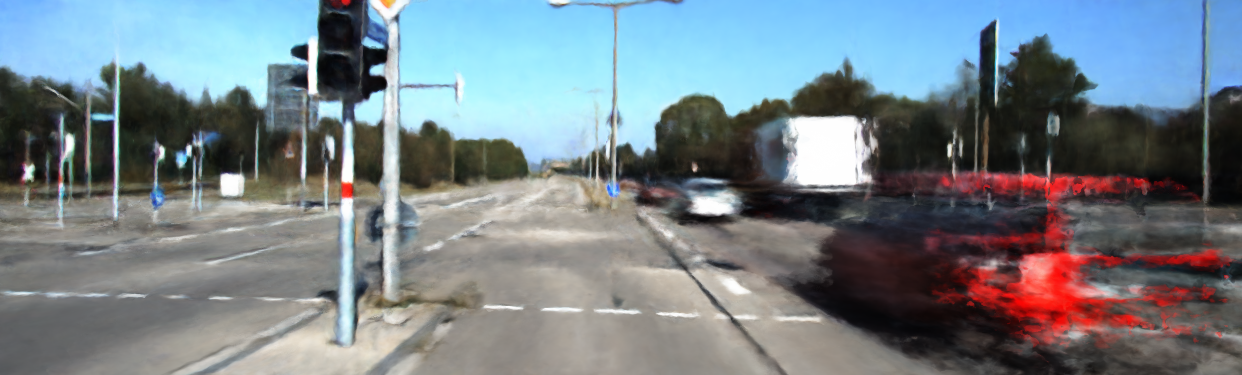}&
		\includegraphics[width=\textwidth/6, trim={400 0 0 0},clip]{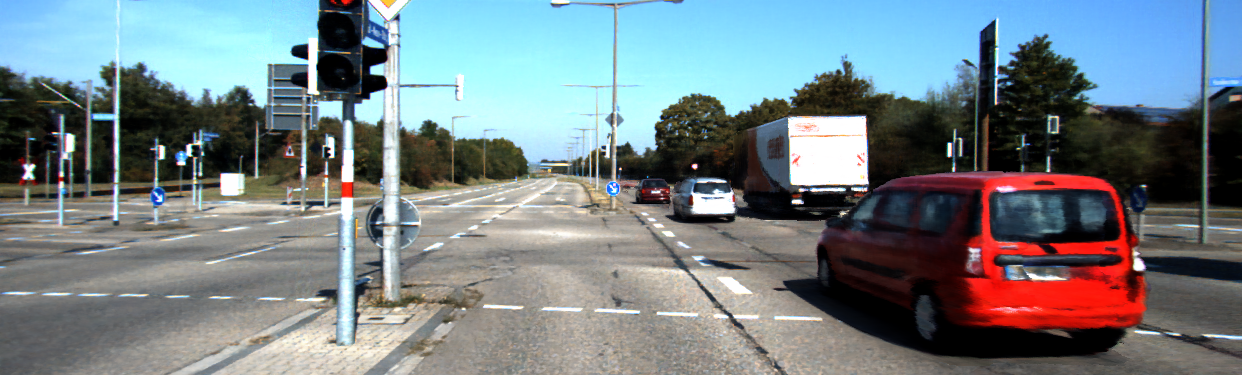}\\
	\end{tabular}\vspace{-6pt}
	\caption{Qualitative results on reconstruction and novel scene arrangements of a scene from the KITTI dataset \cite{geiger2012kittivisonbenchmark} for SRN \cite{sitzmann2019srns}, NeRF \cite{mildenhall2020nerf}, a modified NeRF with a time parameter, and our neural scene graphs. Reconstruction here refers to the reproduction of a frame seen during training, and novel scene arrangements render a scene not seen in the training set. SRN learns to average all frames in the training set. NeRF and the modified variant struggle to learn dynamic parts of the scene adequately. Our neural scene graph method achieves high-quality view synthesis results regardless of this dynamicity, both for dynamic scene reconstruction and novel scene generation.}
	\label{fig:qual_results}
	
	\vspace{-14pt}
\end{figure*}

\subsection{Quantitative Validation}\label{ssec:comparison}
We quantitatively validate our method with comparisons against Scene Representation Networks
(SRNs)~\cite{sitzmann2019srns}, NeRF~\cite{mildenhall2020nerf}, and a modified variant of NeRF on novel scene and reconstruction tasks. For comparison on the method complexity, we refer to the Supplemental Material.

Specifically, we learn neural scene graphs on video sequences of the KITTI data and assess the quality of reconstruction of seen frames using the learned graph, and novel scene compositions. SRNs and NeRF were designed for static scenes and are state-of-the-art approaches for implicit scene representations. To improve the capabilities of NeRF for dynamic scenes, we add a time parameter to the model inputs and denote this approach ``NeRF + time''. The time parameter is positional encoded and concatenated with $\gamma_{\boldsymbol{x}}(\boldsymbol{x})$.  We train both NeRF based methods with the configuration reported for their real forward-facing experiments. Tab.~\ref{tab:quant_results} reports the results which validate that complex dynamic scenes cannot be learned only by adding a time parameter to existing implicit methods.

\vspace{6pt}
\noindent\textbf{Reconstruction Quality}
In Tab.~\ref{tab:quant_results}, we evaluate all methods for scenes from KITTI and present
quantitative results with the PSNR, SSIM\cite{wang2003ssim}, LPIPS \cite{zhang2018perceptual}
metrics. To also evaluate the consistency of the reconstruction for adjacent frames, we evaluate the renderings with two temporal metrics tOF and tLP from TecoGAN~\cite{chu2018tecogan}. The tOF metric examines errors in motion of the reconstructed video comparing the optical flow to the ground truth and for tLP the perceptual distance as the change in LPIPS between consecutive frames is compared.
The proposed method outperforms all baseline methods on all metrics. As shown in the
top row of Fig.~\ref{fig:qual_results}, vanilla SRN and NeRF implicit rendering fail to
accurately reconstruct dynamic scenes. Optimizing SRNs leads to a single static representation
for all frames. For NeRF, even small changes in the camera pose lead to a slightly different viewing direction for spatial points at all frames, and, as such, the method suffers from severe ghosting. NeRF adjusted for temporal changes improves the quality, but still lacks detail and suffers from blurry, uncertain predictions. Significant improvement in the temporal metrics show that our method, which reconstructs each object individually, yields an improved temporal consistency for the reconstructions of the moving objects.  \\

\begin{table}[t]
	\vspace{-0pt}
	\centering
	\resizebox{\columnwidth}{!}{%
	\begin{tabular}[htb]{l|cccc}
		& SRN \cite{sitzmann2019srns}
		& NeRF \cite{mildenhall2020nerf} 
		& NeRF + time
		& Ours \\
		\hline
		\multicolumn{5}{l}{Reconstruction} \\
		\hline\hline
		PSNR $\uparrow$ & 18.83 & 23.34 & 24.18 & \textbf{26.66} \\
		SSIM $\uparrow$  & 0.590  &  0.662 & 0.677 & \textbf{0.806} \\
		LPIPS $\downarrow$ & 0.456 & 0.415 & 0.425 & \textbf{0.186} \\
		tOF $\times 10^{6}$ $\downarrow$ & 1.191 & 1.178 & 1.443 & \textbf{0.765} \\
		tLP $\times 100$ $\downarrow$ & 2.965 & 3.464 & 0.897 & \textbf{0.246} \\

		\hline
		\multicolumn{5}{l}{Novel Composition} \\
		\hline\hline
		PSNR $\uparrow$ & 18.83 & 18.25 & 19.68 & \textbf{25.11} \\
		SSIM $\uparrow$  & 0.590  &  0.594 & 0.593 & \textbf{0.789} \\
		LPIPS $\downarrow$ & 0.456 & 0.442 & 0.473 & \textbf{0.204} \\
	\end{tabular}
	}%
	\caption{We report PSNR, SSIM, LPIPS, tOF and tLP results on scenes from KITTI
  \cite{geiger2012kittivisonbenchmark} for SRN \cite{sitzmann2019srns}, NeRF
  \cite{mildenhall2020nerf}, a modified NeRF variant with an added time input and our neural
  scene graph method. For PSNR and SSIM, higher is better; for LPIPS, tOF and tLP lower is better. Our method outperforms methods designed for static scenes for reconstructing dynamic scenes. For novel compositions, it outperforms existing methods in all image quality metrics.}
	\label{tab:quant_results}

	\vspace{-12pt}
\end{table}

\vspace{-4pt}
\noindent\textbf{Novel Scene Compositions}
To compare the quality of novel scene compositions, we trained all methods on each image
sequence leaving out a single frame from each. In Fig.~\ref{fig:qual_results} (bottom) we report
qualitative results and Tab.~\ref{tab:quant_results} lists the corresponding evaluations. Even
though NeRF and time-aware NeRF are able to recover changes in the scene when they occur with
changing viewing direction, both methods are not able to reason about the scene dynamics. In
contrast, the proposed method is able to adequately synthesize shadows and reflections without
changing viewing direction.


\section{3D Object Detection as Inverse Rendering}\label{ch:detection}

\begin{figure}[t!]
	\centering
	\setlength{\tabcolsep}{1pt}
	\begin{tabular}{cc}
		\includegraphics[width=.49\columnwidth]{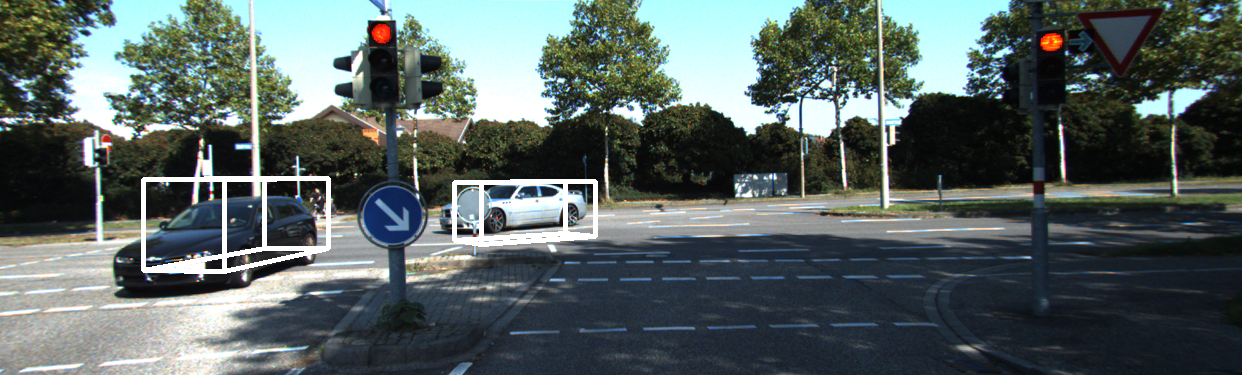}
		\includegraphics[width=.49\columnwidth]{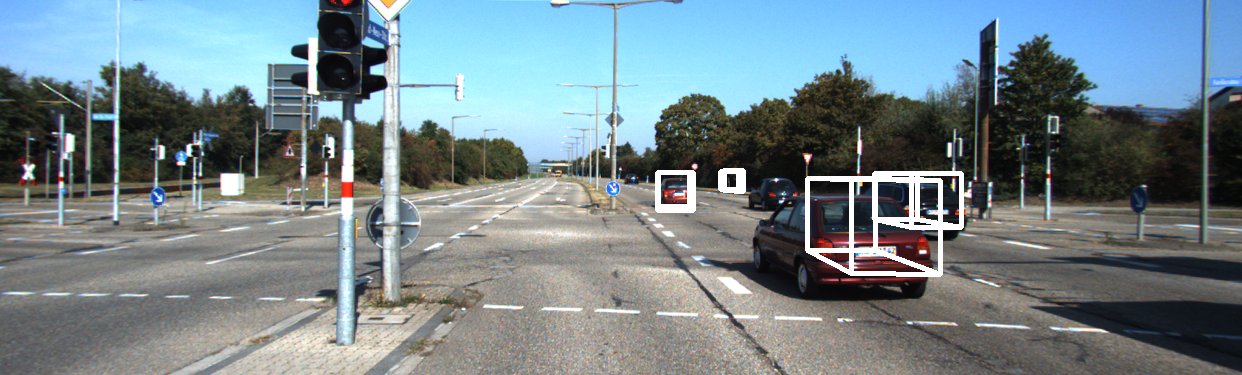}
	\end{tabular}\vspace{-6pt}
	\caption{3D Object detection with neural scene graph rendering. The detected objects and their 3D bounding boxes on the observed images are the result of an optimization over the space of learned scene graphs. The visualized 3D object detections correspond to the scene graph that synthesizes an image with a minimum distance to the observed image.}
	\label{fig:od_example}
	\vspace{-16pt}
\end{figure}

The ability to decompose the scene into multiple objects in scene graphs also allows for improved scene understanding. In this section, we apply
learned scene graphs to the problem of 3D object detection, using the proposed method as a forward model in a single shot learning approach.\\
We formulate the 3D object detection problem as an image synthesis problem over the space of learned scene graphs that best reconstructs a given input image. Specifically, we sample anchor positions in a bird's-eye view plane and optimize over anchor box positions and latent
object codes that minimize the $\ell_1$ image loss between the synthesized image and an observed image. The resulting object detection method is able to find the poses and box dimensions of objects as shown in Fig.~\ref{fig:od_example}.
This application highlights the potential of neural rendering pipelines as a forward rendering model for computer vision tasks, promising additional applications beyond view extrapolation and synthesis.
\section{Discussion and Future Work}
We present the first approach that tackles the challenge of
representing dynamic, multi-object scenes implicitly using neural networks. Using video and annotated tracking
data, our method learns a graph-structured, continuous, spatial representation of
multiple dynamic and static scene elements, automatically decomposing the scene into multiple
independent graph nodes. Comprehensive experiments on simulated and real data achieve
photo-realistic quality previously only attainable with static scenes supervised with many views.
We validate the learned scene graph method by generating
novel arrangements of objects in the scene using the learned graph structure. This approach also
substantially improves on the efficiency of the neural rendering pipeline, which is critical for
learning from video sequences. Building on these contributions, we also present a first method that tackles automotive object detection using a neural rendering approach, validating the potential of the proposed method. \\
Due to the nature of implicit methods, the learned representation quality of the proposed method is bounded by the variation and
amount of training data. In the future, larger view extrapolations might be handled by scene priors learned from large-scale video datasets. We believe that the proposed approach opens up the field of neural rendering for dynamic scenes, and, encouraged by our detection results, may potentially
serve as a method for unsupervised training of computer vision models in the future.





{\small
\bibliographystyle{ieee_fullname}
\bibliography{bib}
}
\end{document}